\documentclass{article} 
\usepackage{iclr2027_conference,times}
\usepackage{pifont}

\usepackage{amsmath,amsfonts,bm}

\def\eqref#1{equation~\ref{#1}}

\def\1{\bm{1}}

\DeclareMathAlphabet{\mathsfit}{\encodingdefault}{\sfdefault}{m}{sl}
\SetMathAlphabet{\mathsfit}{bold}{\encodingdefault}{\sfdefault}{bx}{n}

\usepackage[most]{tcolorbox}
\usepackage{xcolor}
\usepackage{pifont}
\definecolor{echoRose}{HTML}{D59293}
\definecolor{echoRoseDark}{HTML}{A65F62}
\definecolor{softGrid}{HTML}{E2E5E8}
\definecolor{zeroGray}{HTML}{7B8187}
\newcommand{\hlToken}[1]{\textcolor{cyan!60!black}{#1}}
\usepackage{hyperref}
\usepackage{url}
\newcommand{\ourmethod}{{\fontfamily{lmtt}\selectfont \textbf{EchoDistill}}\xspace}
\usepackage[table]{xcolor}
\usepackage{array}

\definecolor{echoBlue}{HTML}{9CC4E6}
\definecolor{echoBlueText}{HTML}{3F6F91}
\usepackage{graphicx}
\usepackage{wrapfig}
\usepackage{capt-of}
\usepackage{booktabs}
\usepackage{multirow}
\usepackage{placeins}
\usepackage{wrapfig}
\usepackage{float}
\usepackage{amssymb}
\usepackage{booktabs}   
\usepackage{multirow}   
\usepackage{graphicx}   
\usepackage{booktabs}      
\usepackage{multirow}      
\usepackage{tabularx}      
\usepackage[table]{xcolor} 
\usepackage{amsmath}       
\usepackage{enumitem}
\usepackage{pgfplots}
\usepackage{tikz}
\usepackage{xcolor}
\pgfplotsset{compat=1.18}
\hypersetup{
    colorlinks=true,
    linkcolor=red,
    citecolor=cyan,
    filecolor=magenta,      
    urlcolor=magenta,
    }
\usepackage{xspace}

\title{EchoDistill: Robust Large Audio Language Models via Noisy-to-Clean Self-Distillation}

\author{
\textbf{
Kaiwen Luo$^{1,*}$,
Chunxi Luo$^{2,*}$,
Liang Lin$^{3,*}$,
Yuxuan Li$^{3}$, 
Zhenhong Zhou$^{1}$,
Junhao Dong$^{1,\dagger}$,}
\\
\textbf{
Yingjie Zhou$^{4}$,
Zhendong Chu$^{5}$
}\\[0.5em]
$^{1}$NTU,
$^{2}$SHU,
$^{3}$UCAS,
$^{4}$SCU,
$^{5}$Squirrel AI
}
\usepackage{tikz}
\usepackage{pgfplots}
\usepgfplotslibrary{groupplots}
\pgfplotsset{compat=1.18}

\usepackage[table]{xcolor}
\definecolor{highlightblue}{rgb}{0.9, 0.95, 1.0}

\iclrfinalcopy
\begin{document}

\maketitle
\thispagestyle{empty}
\fancyhead{}
\renewcommand{\headrulewidth}{0pt}

\begin{abstract}
Large Audio Language Models (LALMs) remain vulnerable to acoustic noise, which can obscure task-relevant evidence and produce unreliable responses. We propose \ourmethod{}, a noisy-to-clean self-distillation framework that uses clean audio as privileged information during post-training. A noisy-input student samples candidate responses reflecting its inference-time behavior, while a frozen copy of the same backbone processes the corresponding clean audio. \ourmethod{} combines masked response-token distillation, task-gated consistency shaping, and teacher-referenced group-relative optimization to align noisy-input generation with clean-conditioned semantics. Only the student is retained at inference time, introducing no additional inference cost. Across three LALM backbones and three audio domains at $-10\,\mathrm{dB}$, \ourmethod{} improves average noisy-input accuracy by 1.63 percentage points over the strongest baseline. On Qwen2.5-Omni, it raises noisy-input accuracy from 59.33\% to 62.94\%, while clean-audio accuracy increases from 76.56\% to 77.56\%. Replacing matched audio with random, shuffled, or silent inputs reduces accuracy by 3.08--6.42 points, confirming that matched acoustic evidence contributes to its predictions. Additional evaluations show improvements on held-out additive noises and external benchmarks, while revealing that these gains do not reliably extend to non-additive distortions. These results demonstrate robust post-training improvements under severe additive noise without sacrificing clean-audio capability across diverse tasks.

\end{abstract}

\section{Introduction}

Large Audio Language Models (LALMs) \cite{chu2024qwen2,tang2024salmonn,ding2025kimi,yang2026moss} have emerged as a vital research direction, leveraging LLMs' advanced representation learning for audio processing and a wide range of applications, including automatic speech recognition \cite{an2025fun,shi2026qwen3}, translation \cite{huang2023speech,du2024cot} and text-to-speech \cite{hu2026qwen3,gong2026moss}. However, audio captured in practical environments is frequently contaminated by additive environmental noise, which distorts the observed waveform and masks task-relevant acoustic cues \cite{song2026stabletoken,zhang2026rsa}. Consequently, the accuracy and reliability of LALM outputs can degrade substantially \cite{yin2024leveraging,kumar2025robustness,luo2026survey,lin2026hidden,luo2026chronosaudio}. Under severe noise, models may misinterpret acoustic evidence and produce unstable responses, leading to semantic drift and hallucinations \cite{chen2025noise, cheng2025omnichat, yin2026focus}.


Existing research on LALM denoising has predominantly focused on inference-time interventions or feature-level modifications \cite{chen2025noise,yin2026focus,chen2026aha,zhang2026see}. These approaches suppress estimated noise or attenuate noise-related representations, but do not teach the model to recover clean-conditioned semantics from corrupted inputs, leaving its predictions vulnerable to semantic drift under severe noise. Moreover, nuisance components and task-relevant acoustic cues are entangled in the learned representation space; aggressive suppression may discard useful evidence and shift acoustic representations away from the distribution expected by the language backbone \cite{song2026stabletoken,zhang2026see}, thereby degrading downstream accuracy or general capabilities. Although methods such as DFL \cite{purwins2019deep} can mitigate certain noise corruptions, a recent study SEE \cite{zhang2026see} shows that such interventions may incur a representation mismatch and downstream degradation.


In this work, we introduce \ourmethod{}, a noisy-to-clean
self-distillation framework for improving the robustness of LALMs \cite{xiaomi2025mimo,yu2026salmonn,lin2026stepaudio} to
severe additive acoustic noise. Our key observation is that paired clean
and noisy recordings preserve the same task-relevant semantics, while
differing in the acoustic corruption applied to the observation.
Consequently, under the same response context, a change in the model's
next-token preference between the two recordings directly reveals the
semantic decision altered by noise \cite{lopez2025robustness,ma2025towards}. Based on this relationship,
\ourmethod{} uses the clean-conditioned prediction as a reference for
what the model would have predicted from the same content without
corruption, and corrects the corresponding noisy-conditioned token
preference before such local errors accumulate into sequence-level
semantic drift. Unlike existing methods that estimate and suppress noise
in the waveform or representation space, this approach directly corrects
the semantic consequence of noise without assuming that interference can
be cleanly separated from task-relevant acoustic evidence.

To implement this principle, \ourmethod{} exploits an information
asymmetry available during post-training. The student receives noisy
audio and generates candidate responses, whereas a frozen copy of the
same backbone processes the corresponding clean audio. For each
student-generated prefix, we align the paired next-token distributions
through masked response-token distillation. Noisy-input rollouts expose
the decisions encountered at inference time, while a teacher-referenced
group-relative clipped objective reinforces trajectories corrected
toward clean-conditioned semantics. A task-positive consistency term
further prevents agreement with an incorrect teacher prediction from
being rewarded. Clean audio and the teacher branch are used only during
post-training; at inference time, \ourmethod{} retains the original
single-model architecture without additional cost.


As shown in Figure~\ref{fig:taskwise_comparison}, we evaluate \ourmethod{} under severe additive acoustic noise at $-10\,\mathrm{dB}$ across three representative LALMs: Qwen2.5-Omni \cite{Qwen2.5-Omni-Report}, MiniCPM-o \cite{yao2024minicpm}, and Step-Audio-2-mini \cite{wu2025step} and three audio domains: Music, Sound, and Speech \cite{sakshi2025mmau}. We use noisy-input accuracy as the primary metric of task correctness and report Generation Success Rate (GSR) as an auxiliary measure of behavioral consistency. Averaged over the nine backbone--domain settings, \ourmethod{} improves noisy-input accuracy by 1.63 percentage points and GSR by 4.18 percentage points over the strongest corresponding baselines. These improvements show that \ourmethod{} not only stabilizes clean--noisy behavior but also converts this stability into more accurate predictions. Our contributions are as follows:




\begin{itemize}[
    leftmargin=*,
    topsep=0.25em,
    itemsep=0.45em,
    parsep=0pt,
    partopsep=0pt
]
\item \ding{182}We formulate LALM robustness post-training as \emph{on-trajectory semantic correction with privileged clean audio}. For each noisy-student response prefix, paired clean and noisy next-token distributions reveal how acoustic corruption shifts semantic decisions, rather than relying on feature suppression or offline teacher responses.
\item \ding{183}We propose \ourmethod{}, a closed-loop noisy-to-clean self-distillation framework. Noisy rollouts expose inference-time errors, masked token-level distillation provides clean-conditioned corrections along the same trajectories, and task-positive consistency shaping prevents rewarding incorrect teacher agreement. A teacher-referenced group-relative objective reinforces the corrected trajectories without additional inference cost.
\item \ding{184}We conduct a correctness-centered evaluation across three LALM backbones and three audio domains. Clean-audio retention, controlled baselines, SNR sweeps, held-out noises, cross-benchmark transfer, and audio interventions demonstrate improvements under severe additive noise while identifying the generalization boundary under non-additive distortions.
\end{itemize}

\raggedbottom
\setlength{\dbltextfloatsep}{8pt}
\setlength{\dblfloatsep}{8pt}

\begin{figure*}[!t]
    \centering

    \definecolor{qwen3b}{RGB}{70,130,230}
    \definecolor{qwen7b}{RGB}{245,140,30}
    \definecolor{gptaudio}{RGB}{75,165,75}
    \definecolor{student}{RGB}{250,40,35}

    {\small\bfseries
    \setlength{\tabcolsep}{9pt}
    \renewcommand{\arraystretch}{1.0}
    \begin{tabular}{cccc}
        \tikz[baseline=-0.5ex]{
            \draw[
                draw=qwen3b,
                fill=qwen3b,
                fill opacity=0.42,
                line width=0.95pt
            ] (0,0) rectangle (0.34,0.16);
        }
        Qwen2.5-Omni-3B
        &
        \tikz[baseline=-0.5ex]{
            \draw[
                draw=qwen7b,
                fill=qwen7b,
                fill opacity=0.42,
                line width=0.95pt
            ] (0,0) rectangle (0.34,0.16);
        }
        Qwen2.5-Omni-7B
        &
        \tikz[baseline=-0.5ex]{
            \draw[
                draw=gptaudio,
                fill=gptaudio,
                fill opacity=0.42,
                line width=0.95pt
            ] (0,0) rectangle (0.34,0.16);
        }
        GPT-audio
        &
        \tikz[baseline=-0.5ex]{
            \draw[
                draw=student,
                fill=student,
                fill opacity=0.48,
                line width=1.10pt
            ] (0,0) rectangle (0.34,0.16);
        }
        Student from Qwen
    \end{tabular}
    }

    \vspace{0.6mm}

    \resizebox{0.98\textwidth}{!}{%
    \begin{tikzpicture}

        \pgfplotsset{
            qwen3bbar/.style={
                draw=qwen3b,
                fill=qwen3b,
                fill opacity=0.42,
                line width=0.95pt,
                every node near coord/.append style={xshift=-4.5pt}
            },
            qwen7bbar/.style={
                draw=qwen7b,
                fill=qwen7b,
                fill opacity=0.42,
                line width=0.95pt,
                every node near coord/.append style={xshift=-1.5pt}
            },
            gptaudiobar/.style={
                draw=gptaudio,
                fill=gptaudio,
                fill opacity=0.42,
                line width=0.95pt,
                every node near coord/.append style={xshift=1.5pt}
            },
            studentbar/.style={
                draw=student,
                fill=student,
                fill opacity=0.48,
                line width=1.10pt,
                every node near coord/.append style={xshift=4.5pt}
            }
        }

        \begin{groupplot}[
            group style={
                group size=3 by 1,
                horizontal sep=1.0cm
            },
            ybar,
            ymin=50,
            ymax=80,
            ytick={50,60,70,80},
            symbolic x coords={Noisy,GSR},
            xtick=data,
            xticklabels={
                {Noisy Acc. (\%) $\uparrow$},
                {GSR (\%) $\uparrow$}
            },
            enlarge x limits=0.50,
            width=5.95cm,
            height=3.55cm,
            axis x line*=bottom,
            axis y line*=left,
            axis line style={black!65,line width=0.5pt},
            tick style={black!65,line width=0.5pt},
            tick align=outside,
            ymajorgrids=true,
            major grid style={
                black!18,
                dashed,
                line width=0.4pt
            },
            clip=false,
            nodes near coords,
            every node near coord/.append style={
                font=\scriptsize\bfseries,
                text=black!60,
                yshift=1.4pt
            },
            xticklabel style={
                font=\small\bfseries,
                align=center,
                yshift=1pt
            },
            yticklabel style={font=\small},
            title style={
                font=\large\bfseries,
                yshift=-2pt
            },
            ylabel style={
                font=\small\bfseries,
                yshift=-2pt
            }
        ]

        \nextgroupplot[
            title={(a) Music},
            ylabel={Performance (\%)}
        ]
        \addplot[qwen3bbar]
            coordinates {(Noisy,50.8) (GSR,66.4)};
        \addplot[qwen7bbar]
            coordinates {(Noisy,59.0) (GSR,74.0)};
        \addplot[gptaudiobar]
            coordinates {(Noisy,51.4) (GSR,67.5)};
        \addplot[studentbar]
            coordinates {(Noisy,59.2) (GSR,75.1)};

        \nextgroupplot[
            title={(b) Sound}
        ]
        \addplot[qwen3bbar]
            coordinates {(Noisy,60.8) (GSR,67.1)};
        \addplot[qwen7bbar]
            coordinates {(Noisy,58.2) (GSR,65.5)};
        \addplot[gptaudiobar]
            coordinates {(Noisy,52.2) (GSR,68.2)};
        \addplot[studentbar]
            coordinates {(Noisy,66.8) (GSR,75.3)};

        \nextgroupplot[
            title={(c) Speech}
        ]
        \addplot[qwen3bbar]
            coordinates {(Noisy,54.5) (GSR,69.8)};
        \addplot[qwen7bbar]
            coordinates {(Noisy,60.8) (GSR,74.5)};
        \addplot[gptaudiobar]
            coordinates {(Noisy,53.7) (GSR,75.4)};
        \addplot[studentbar]
            coordinates {(Noisy,62.9) (GSR,78.0)};

        \end{groupplot}
    \end{tikzpicture}%
    }

    \vspace{-1.5mm}
    \caption{%
        Task-wise noisy-input accuracy and Generation Success Rate (GSR)
        under severe additive noise at $-10\,\mathrm{dB}$.
        Accuracy measures task correctness, whereas GSR measures
        clean--noisy prediction consistency. EchoDistill improves both
        metrics across the Music, Sound, and Speech domains.
    }
    \label{fig:taskwise_comparison}
    \vspace{-1.0em}
\end{figure*}

\section{Related Work and Background}
\label{sec:related_background}

\subsection{Related Work}

\paragraph{Noise robustness for speech and LALMs.}
Traditional robust speech systems use data augmentation, speech enhancement,
and source separation to reduce acoustic mismatch
\citep{chen2025noise,yin2026focus,song2026stabletoken,zhang2026see}.
Recent LALM methods further introduce task-conditioned enhancement or
representation-level noise suppression
\citep{yin2024leveraging,chen2025noise,yin2026focus,zhang2026see}.
These approaches modify waveforms or acoustic representations and may remove
task-relevant cues when signal and interference overlap
\citep{purwins2019deep,zhang2026see}.
Noise-augmented SFT instead provides answer-level supervision but cannot
localize corruption-induced errors along noisy-input trajectories.
\ourmethod{} addresses this limitation through on-trajectory semantic
correction without adding an inference-time enhancement module.

\paragraph{Knowledge distillation and on-policy self-distillation.}
Knowledge distillation transfers predictive distributions from a teacher to
a student \citep{hinton2015distilling}.
Offline distillation may suffer from a mismatch between fixed training
sequences and student-generated trajectories \citep{ yang2026learning},
whereas on-policy distillation queries the teacher on the student's own
outputs \citep{agarwal2024policy,zhao2026self,ko2026scaling}.
Unlike conventional teacher imitation, \ourmethod{} evaluates each
student-generated prefix under paired clean and noisy audio. Their next-token
discrepancy estimates the semantic shift caused by acoustic corruption and
provides a trajectory-specific correction signal. Extended discussion is
provided in Appendix~\ref{app:related_work}.

\subsection{Background}

\paragraph{Self-distillation.}
Let $p_{\phi}$ and $p_{\theta}$ denote the teacher and student distributions.
Output-level self-distillation is commonly formulated as
\begin{equation}
\mathcal{L}_{\mathrm{SD}}
=
\frac{1}{T}\sum_{t=1}^{T}
D_{\mathrm{KL}}
\left(
p_{\phi}(\cdot\mid x,y_{<t})
\,\Vert\,
p_{\theta}(\cdot\mid x,y_{<t})
\right).
\end{equation}
In \ourmethod{}, the two distributions share the same student-generated
prefix but are conditioned on clean and noisy audio, respectively.

\paragraph{Large Audio Language Models.}
LALMs couple an audio encoder with a language backbone
\citep{huang2024audiogpt,xu2025qwen3,luo2026survey}.
Given audio $a$ and prompt $x$, their generation process can be written as
\begin{equation}
\pi(y\mid x,a)
=
\prod_{t=1}^{T}
\pi\!\left(y_t\mid x,g(E(a)),y_{<t}\right),
\end{equation}
where $E(\cdot)$ is the audio encoder and $g(\cdot)$ maps acoustic features
into the language model. Noise can corrupt this conditioning representation
and consequently shift the generated response.

\paragraph{Group Relative Policy Optimization.}
Group Relative Policy Optimization (GRPO)
\citep{shao2024deepseekmath} samples a group of responses
$\{y^{(k)}\}_{k=1}^{K}$ for each input and estimates their relative
advantages by normalizing rewards within the group:
\begin{equation}
A^{(k)}
=
\frac{r^{(k)}-\mu}{\sigma+\varepsilon},
\end{equation}
where $\mu$ and $\sigma$ denote the group reward mean and standard
deviation. GRPO then optimizes a clipped importance-ratio objective
between the current and old policies. However, because the same sequence-level advantage is
assigned to all response tokens, standard GRPO cannot identify which
token-level decisions are altered by acoustic corruption.

\section{Methodology}
\label{sec:method}

\subsection{Problem Formulation}

We consider post-training a LALM using paired clean and noisy recordings that convey the same semantic content. Let
\begin{equation}
\mathcal{D}
=
\left\{
\left(
x_i,
a_i^{\mathrm{n}},
a_i^{\mathrm{c}},
y_i^\star,
\mathcal{C}_i
\right)
\right\}_{i=1}^{N},
\end{equation}
where $x_i$ is the textual instruction, $a_i^{\mathrm{n}}$ and
$a_i^{\mathrm{c}}$ are the noisy and clean versions of the same audio,
$y_i^\star$ is the ground-truth answer, and $\mathcal{C}_i$ denotes the
candidate-answer set when applicable.

We initialize a trainable student policy $\pi_\theta$ and a frozen teacher
$\pi_\phi$ from the same pretrained model. The student receives
$(x_i,a_i^{\mathrm{n}})$, whereas the teacher receives
$(x_i,a_i^{\mathrm{c}})$. The clean audio therefore serves as privileged
information during post-training rather than as an additional inference-time
input. Our objective is to improve
$\pi_\theta(y\mid x_i,a_i^{\mathrm{n}})$ while preserving its performance on
uncorrupted audio. At inference time, only the trained student and the observed
audio are required.

\subsection{\ourmethod{} Framework}
\label{sec:framework}
\ourmethod{} combines three complementary components:
noisy-input rollouts that expose the student's inference-time behavior,
token-level noisy-to-clean distillation that supplies dense clean-conditioned
supervision, and task-gated consistency shaping that differentiates sampled
responses using their candidate-specific clean--noisy discrepancy.

\begin{figure*}[t]
    \centering
    \includegraphics[width=\textwidth]{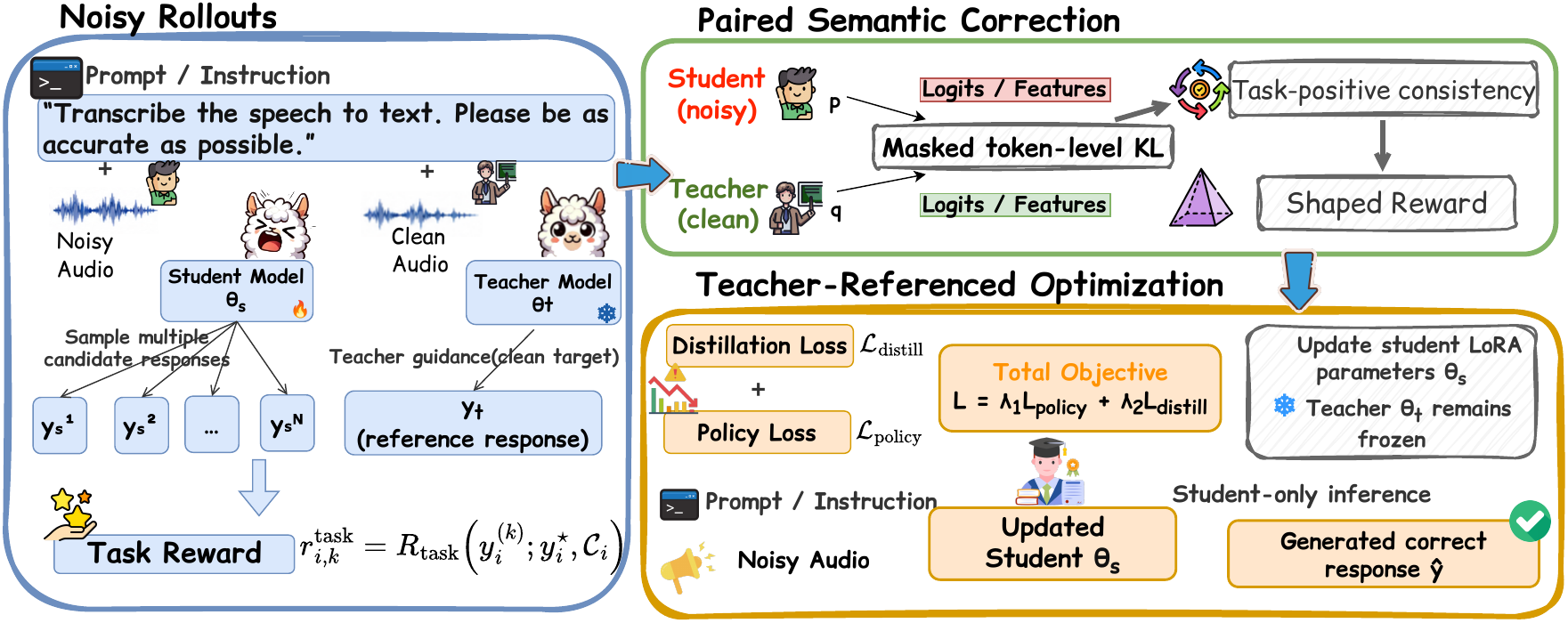}
    \caption{
        \textbf{Overview of \ourmethod{}.}
        The noisy-input student first samples candidate responses. The student and frozen clean-audio teacher produce
paired token distributions, whose masked discrepancy provides token-level
correction and a task-gated consistency bonus. A teacher-referenced
group-relative objective updates only the student, which is used alone at
inference time.
    }
    \label{fig:echodistill_pipeline}
\end{figure*}

\paragraph{Noisy-input rollouts.}

At the beginning of each update, we retain a fixed rollout policy
$\pi_{\theta_{\mathrm{old}}}$ and sample $K$ candidate responses:

\begin{equation}
y_i^{(k)}
\sim
\pi_{\theta_{\mathrm{old}}}
(\cdot\mid x_i,a_i^n),
\qquad
r_{i,k}^{\mathrm{task}}
=
R_{\mathrm{task}}
\left(y_i^{(k)};y_i^\star,\mathcal{C}_i\right),
\quad k=1,\ldots,K.
\end{equation}

\paragraph{Candidate-specific noisy-to-clean distillation.}

For each sampled response, we evaluate the frozen teacher and trainable student
along the same textual prefix but under different acoustic conditions. Here, $m_{i,k,t}$ selects assistant-response tokens and excludes prompt tokens,
audio placeholders, and dialogue-template tokens. The stop-gradient operator
$\operatorname{sg}[\cdot]$ prevents gradients from flowing through the frozen
teacher. The instance-level distillation objective is
\begin{equation}
\begin{aligned}
D_{i,k}(\theta)
&=
\frac{
\sum_t m_{i,k,t}
D_{\mathrm{KL}}
\left(
\operatorname{sg}[q_{i,k,t}]
\Vert
p_{i,k,t}^{\theta}
\right)}
{\sum_t m_{i,k,t}},
\qquad
\mathcal{L}_{\mathrm{distill}}^{(i)}=
\frac{1}{K}\sum_{k=1}^{K}D_{i,k}(\theta).
\end{aligned}
\end{equation}

Unlike answer-level imitation, this objective transfers the teacher's complete
next-token distribution along student-generated trajectories.

\paragraph{Task-gated clean--noisy consistency shaping.}

We use the candidate-specific discrepancy to augment the task reward. To keep
the reward fixed within an optimization update, the score is evaluated using
the rollout policy and detached from gradient computation:
\begin{equation}
s_{i,k}
=
\operatorname{sg}
\left[
\exp
\left(
-D_{i,k}(\theta_{\mathrm{old}})
\right)
\right].
\end{equation}
The shaped reward is
\begin{equation}
\bar{r}_{i,k}
=
\operatorname{clip}
\left(
r_{i,k}^{\mathrm{task}}
+
\beta b_{i,k}s_{i,k},
-1,2
\right),
\end{equation}
where $\beta$ controls the strength of the consistency bonus. The correctness
gate $b_{i,k}$ prevents an incorrect response from receiving a bonus merely
because its noisy- and clean-conditioned distributions agree. Among verified
correct responses, candidate-specific scores allow different trajectories to
receive different bonuses.

We refer to $s_{i,k}$ as a clean--noisy consistency score rather than a direct
measure of acoustic grounding. Distributional agreement alone does not prove
that a prediction is causally grounded in the audio; this distinction is
examined separately through controlled audio interventions.

\subsection{Teacher-Referenced Group-Relative Optimization}
\label{sec:optimization}

We compute the mean response-token log-likelihoods of each sampled candidate
under the noisy student and clean-conditioned teacher:
\begin{equation}
\ell_{i,k}^{z}(\psi)
=
\frac{
\sum_{t=1}^{T_{i,k}}
m_{i,k,t}
\log \pi_\psi
\left(
y_{i,t}^{(k)}
\mid
x_i,a_i^{z},y_{i,<t}^{(k)}
\right)
}{
\sum_{t=1}^{T_{i,k}}m_{i,k,t}
},
\quad
(z,\psi)\in
\left\{
(\mathrm{n},\theta),(\mathrm{c},\phi)
\right\}.
\label{eq:paired_log_likelihoods}
\end{equation}

Their difference defines a teacher-referenced likelihood weight:
\begin{equation}
w_{i,k}^{\mathrm{TCR}}(\theta)
=
\exp
\left(
\ell_{i,k}^{\mathrm{n}}(\theta)
-
\operatorname{sg}
\left[
\ell_{i,k}^{\mathrm{c}}(\phi)
\right]
\right).
\end{equation}
We optimize the following teacher-referenced group-relative clipped surrogate:
\begin{equation}
\mathcal{L}_{\mathrm{policy}}^{(i)}
=
-\frac{1}{K}
\sum_{k=1}^{K}
\min
\left(
w_{i,k}^{\mathrm{TCR}}A_{i,k},
\operatorname{clip}
\left(
w_{i,k}^{\mathrm{TCR}},
1-\epsilon,
1+\epsilon
\right)
A_{i,k}
\right).
\end{equation}

Because the numerator and denominator of
$w_{i,k}^{\mathrm{TCR}}$ are conditioned on different acoustic inputs, this
quantity is not an importance-sampling ratio between a current and an old
policy. Instead, it is a response-dependent clean-reference weight: the
clean-conditioned likelihood provides a semantic reference scale, while
clipping limits the influence of extreme likelihood differences. Accordingly,
we describe this component as a teacher-referenced clipped surrogate rather
than standard GRPO.

For a minibatch of size $B$, the complete training objective is
\begin{equation}
\mathcal{L}_{\mathrm{total}}
=
\frac{1}{B}
\sum_{i=1}^{B}
\left[
\lambda_{\mathrm{policy}}
\mathcal{L}_{\mathrm{policy}}^{(i)}
+
\lambda_{\mathrm{distill}}
\mathcal{L}_{\mathrm{distill}}^{(i)}
\right],
\end{equation}
where $\lambda_{\mathrm{policy}}$ and $\lambda_{\mathrm{distill}}$ balance
group-relative optimization and token-level distillation. Only $\theta$ is
updated; $\phi$ remains frozen throughout post-training.

\paragraph{Inference.}

After post-training, the teacher branch, clean audio, rollout sampling, and
reward computation are discarded. Given an observed audio input $a$, the model
generates
\begin{equation}
y
\sim
\pi_\theta(\cdot\mid x,a).
\end{equation}

\section{Experiments}
\label{experiments}

We evaluate \ourmethod{} through three research questions:
\begin{itemize}[leftmargin=*]
    \item \textbf{RQ1:}
    Does \ourmethod{} outperform controlled baselines without degrading clean-audio performance, and what do its ablations reveal?

    \item \textbf{RQ2:}
    How well does \ourmethod{} generalize across noise conditions and benchmarks? Under which corruption conditions does the method cease to provide gains?

    \item \textbf{RQ3:}
    To what extent do the predictions of \ourmethod{} depend on acoustic evidence matched to the question, rather than clean--noisy output agreement or input-independent priors?
\end{itemize}

\begin{table*}[t] \footnotesize
    \centering
    \renewcommand{\arraystretch}{1.00}
    \setlength{\tabcolsep}{4pt}
    \setlength{\aboverulesep}{3pt}
    \setlength{\belowrulesep}{3pt}
    \caption{Main results under severe additive noise
    ($\mathrm{SNR}=-10\,\mathrm{dB}$). \texttt{Acc} denotes noisy-input
    task accuracy and is our primary metric; \texttt{Noisy} and \texttt{GSR}
    are auxiliary clean--noisy agreement diagnostics. Red and blue values
    denote percentage-point increases ($\uparrow$) and decreases
    ($\downarrow$) relative to the corresponding Initial model. The best and
    second-best results are shown in \textbf{bold} and
    \underline{underlined}.}
    \label{tab:EchoDistill_Full_Metrics}
    \resizebox{\textwidth}{!}{%
    \begin{tabular}{@{}l|l|ccc|ccc|ccc@{}}
    \toprule
    \multirow{2}{*}{\textbf{Domain}} & \multirow{2}{*}{\textbf{Method}} & \multicolumn{3}{c|}{\textbf{\textit{Step-Audio-2-mini}}} & \multicolumn{3}{c|}{\textbf{\textit{Qwen2.5-Omni}}} & \multicolumn{3}{c}{\textbf{\textit{MiniCPM-o-2.6}}} \\ \cmidrule(l){3-11}
     &  & \textbf{Noisy} & \textbf{Acc} & \textbf{GSR} & \textbf{Noisy} & \textbf{Acc} & \textbf{GSR} & \textbf{Noisy} & \textbf{Acc} & \textbf{GSR} \\ \midrule

    \multirow{6}{*}{\textbf{Music}}
     & \textbf{Initial} & \textemdash & 59.17 & 78.67 & \textemdash & 59.00 & 74.00 & \textemdash & 63.63 & 85.15 \\
     & \textbf{WT} & \underline{75.08} & 59.08 \textcolor{blue!80!black}{\scriptsize $\downarrow_{0.09}$} & 77.33 \textcolor{blue!80!black}{\scriptsize $\downarrow_{1.34}$} & 71.33 & \textbf{59.33 \textcolor{red}{\scriptsize $\uparrow_{0.33}$}} & 74.16 \textcolor{red}{\scriptsize $\uparrow_{0.16}$} & 85.10 & 64.06 \textcolor{red}{\scriptsize $\uparrow_{0.43}$} & 86.95 \textcolor{red}{\scriptsize $\uparrow_{1.80}$} \\
     & \textbf{STFT} & \textbf{78.16} & \textbf{60.58 \textcolor{red}{\scriptsize $\uparrow_{1.41}$}} & \underline{79.00 \textcolor{red}{\scriptsize $\uparrow_{0.33}$}} & 73.91 & 58.75 \textcolor{blue!80!black}{\scriptsize $\downarrow_{0.25}$} & \textbf{75.41 \textcolor{red}{\scriptsize $\uparrow_{1.41}$}} & \textbf{86.49} & 64.55 \textcolor{red}{\scriptsize $\uparrow_{0.92}$} & \underline{87.15 \textcolor{red}{\scriptsize $\uparrow_{2.00}$}} \\
     & \textbf{DFL} & 74.91 & 59.08 \textcolor{blue!80!black}{\scriptsize $\downarrow_{0.09}$} & 75.66 \textcolor{blue!80!black}{\scriptsize $\downarrow_{3.01}$} & 72.75 & 56.58 \textcolor{blue!80!black}{\scriptsize $\downarrow_{2.42}$} & 74.50 \textcolor{red}{\scriptsize $\uparrow_{0.50}$} & 85.40 & \textbf{64.82 \textcolor{red}{\scriptsize $\uparrow_{1.19}$}} & 86.57 \textcolor{red}{\scriptsize $\uparrow_{1.42}$} \\
     & \textbf{SEEN} & 70.75 & 57.00 \textcolor{blue!80!black}{\scriptsize $\downarrow_{2.17}$} & 77.08 \textcolor{blue!80!black}{\scriptsize $\downarrow_{1.59}$} & \underline{74.08} & 58.91 \textcolor{blue!80!black}{\scriptsize $\downarrow_{0.09}$} & 73.91 \textcolor{blue!80!black}{\scriptsize $\downarrow_{0.09}$} & 85.47 & 63.63 \textcolor{blue!80!black}{\scriptsize $\downarrow_{0.00}$} & 85.05 \textcolor{blue!80!black}{\scriptsize $\downarrow_{0.10}$} \\
    \rowcolor{highlightblue} & \textbf{EchoDistill} & \textbf{78.16} & \underline{59.75 \textcolor{red}{\scriptsize $\uparrow_{0.58}$}} & \textbf{86.00 \textcolor{red}{\scriptsize $\uparrow_{7.33}$}} & \textbf{74.58} & \underline{59.17 \textcolor{red}{\scriptsize $\uparrow_{0.17}$}} & \underline{75.08 \textcolor{red}{\scriptsize $\uparrow_{1.08}$}} & \underline{86.32} & \underline{64.69 \textcolor{red}{\scriptsize $\uparrow_{1.06}$}} & \textbf{87.24 \textcolor{red}{\scriptsize $\uparrow_{2.09}$}} \\
    \midrule

    \multirow{6}{*}{\textbf{Sound}}
     & \textbf{Initial} & \textemdash & 63.50 & 70.42 & \textemdash & 58.17 & 65.50 & \textemdash & 61.92 & 77.76 \\
     & \textbf{WT} & 69.00 & 63.33 \textcolor{blue!80!black}{\scriptsize $\downarrow_{0.17}$} & 70.50 \textcolor{red}{\scriptsize $\uparrow_{0.08}$} & 61.58 & 56.66 \textcolor{blue!80!black}{\scriptsize $\downarrow_{1.51}$} & 63.41 \textcolor{blue!80!black}{\scriptsize $\downarrow_{2.09}$} & 75.27 & 60.86 \textcolor{blue!80!black}{\scriptsize $\downarrow_{1.06}$} & 77.66 \textcolor{blue!80!black}{\scriptsize $\downarrow_{0.10}$} \\
     & \textbf{DFL} & \underline{71.50} & 63.50 \textcolor{blue!80!black}{\scriptsize $\downarrow_{0.00}$} & 74.58 \textcolor{red}{\scriptsize $\uparrow_{4.16}$} & 64.66 & 59.75 \textcolor{red}{\scriptsize $\uparrow_{1.58}$} & 67.41 \textcolor{red}{\scriptsize $\uparrow_{1.91}$} & \textbf{78.72} & 61.55 \textcolor{blue!80!black}{\scriptsize $\downarrow_{0.37}$} & \underline{80.31 \textcolor{red}{\scriptsize $\uparrow_{2.55}$}} \\
     & \textbf{STFT} & 70.33 & \textbf{63.75 \textcolor{red}{\scriptsize $\uparrow_{0.25}$}} & 69.91 \textcolor{blue!80!black}{\scriptsize $\downarrow_{0.51}$} & \underline{67.75} & \underline{60.75 \textcolor{red}{\scriptsize $\uparrow_{2.58}$}} & \underline{68.25 \textcolor{red}{\scriptsize $\uparrow_{2.75}$}} & 77.11 & 60.93 \textcolor{blue!80!black}{\scriptsize $\downarrow_{0.99}$} & 77.78 \textcolor{red}{\scriptsize $\uparrow_{0.02}$} \\
     & \textbf{SEEN} & 67.25 & 59.83 \textcolor{blue!80!black}{\scriptsize $\downarrow_{3.67}$} & \underline{77.16 \textcolor{red}{\scriptsize $\uparrow_{6.74}$}} & 66.25 & 59.33 \textcolor{red}{\scriptsize $\uparrow_{1.16}$} & 66.42 \textcolor{red}{\scriptsize $\uparrow_{0.92}$} & \underline{77.20} & \underline{61.56 \textcolor{blue!80!black}{\scriptsize $\downarrow_{0.36}$}} & 78.04 \textcolor{red}{\scriptsize $\uparrow_{0.28}$} \\
    \rowcolor{highlightblue} & \textbf{EchoDistill} & \textbf{71.75} & \underline{63.67 \textcolor{red}{\scriptsize $\uparrow_{0.17}$}} & \textbf{80.08 \textcolor{red}{\scriptsize $\uparrow_{9.66}$}} & \textbf{72.83} & \textbf{66.83 \textcolor{red}{\scriptsize $\uparrow_{8.66}$}} & \textbf{75.33 \textcolor{red}{\scriptsize $\uparrow_{9.83}$}} & 77.14 & \textbf{62.18 \textcolor{red}{\scriptsize $\uparrow_{0.26}$}} & \textbf{80.42 \textcolor{red}{\scriptsize $\uparrow_{2.66}$}} \\
    \midrule

    \multirow{6}{*}{\textbf{Speech}}
     & \textbf{Initial} & \textemdash & 55.42 & 69.15 & \textemdash & 60.80 & 74.56 & \textemdash & 57.27 & 74.64 \\
     & \textbf{WT} & 58.07 & 46.62 \textcolor{blue!80!black}{\scriptsize $\downarrow_{8.80}$} & 59.49 \textcolor{blue!80!black}{\scriptsize $\downarrow_{9.66}$} & 66.52 & 53.21 \textcolor{blue!80!black}{\scriptsize $\downarrow_{7.59}$} & 66.52 \textcolor{blue!80!black}{\scriptsize $\downarrow_{8.04}$} & 68.16 & 50.82 \textcolor{blue!80!black}{\scriptsize $\downarrow_{6.45}$} & 69.43 \textcolor{blue!80!black}{\scriptsize $\downarrow_{5.21}$} \\
     & \textbf{DFL} & 65.55 & 51.83 \textcolor{blue!80!black}{\scriptsize $\downarrow_{3.59}$} & 64.88 \textcolor{blue!80!black}{\scriptsize $\downarrow_{4.27}$} & 72.65 & 58.46 \textcolor{blue!80!black}{\scriptsize $\downarrow_{2.34}$} & 72.15 \textcolor{blue!80!black}{\scriptsize $\downarrow_{2.41}$} & 72.66 & 55.17 \textcolor{blue!80!black}{\scriptsize $\downarrow_{2.10}$} & \underline{75.31 \textcolor{red}{\scriptsize $\uparrow_{0.67}$}} \\
     & \textbf{STFT} & \underline{68.22} & \underline{54.25 \textcolor{blue!80!black}{\scriptsize $\downarrow_{1.17}$}} & 68.47 \textcolor{blue!80!black}{\scriptsize $\downarrow_{0.68}$} & 74.30 & 59.09 \textcolor{blue!80!black}{\scriptsize $\downarrow_{1.71}$} & 73.97 \textcolor{blue!80!black}{\scriptsize $\downarrow_{0.59}$} & 71.80 & 55.62 \textcolor{blue!80!black}{\scriptsize $\downarrow_{1.65}$} & 75.11 \textcolor{red}{\scriptsize $\uparrow_{0.47}$} \\
     & \textbf{SEEN} & 66.63 & 51.34 \textcolor{blue!80!black}{\scriptsize $\downarrow_{4.08}$} & \underline{70.48 \textcolor{red}{\scriptsize $\uparrow_{1.33}$}} & \underline{75.14} & \underline{59.79 \textcolor{blue!80!black}{\scriptsize $\downarrow_{1.01}$}} & \underline{74.47 \textcolor{blue!80!black}{\scriptsize $\downarrow_{0.09}$}} & \underline{73.07} & \underline{57.91 \textcolor{red}{\scriptsize $\uparrow_{0.64}$}} & 73.27 \textcolor{blue!80!black}{\scriptsize $\downarrow_{1.37}$} \\
    \rowcolor{highlightblue} & \textbf{EchoDistill} & \textbf{71.04} & \textbf{55.29 \textcolor{blue!80!black}{\scriptsize $\downarrow_{0.13}$}} & \textbf{74.89 \textcolor{red}{\scriptsize $\uparrow_{5.74}$}} & \textbf{78.67} & \textbf{62.89 \textcolor{red}{\scriptsize $\uparrow_{2.09}$}} & \textbf{78.01 \textcolor{red}{\scriptsize $\uparrow_{3.45}$}} & \textbf{73.75} & \textbf{58.49 \textcolor{red}{\scriptsize $\uparrow_{1.22}$}} & \textbf{76.46 \textcolor{red}{\scriptsize $\uparrow_{1.82}$}} \\
    \midrule

    \multirow{6}{*}{\textbf{Average}}
     & \textbf{Initial} & \textemdash & 59.36 & 72.75 & \textemdash & 59.32 & 71.35 & \textemdash & 60.94 & 79.18 \\
     & \textbf{WT} & 67.38 & 56.34 \textcolor{blue!80!black}{\scriptsize $\downarrow_{3.02}$} & 69.11 \textcolor{blue!80!black}{\scriptsize $\downarrow_{3.64}$} & 66.48 & 56.40 \textcolor{blue!80!black}{\scriptsize $\downarrow_{2.92}$} & 68.03 \textcolor{blue!80!black}{\scriptsize $\downarrow_{3.32}$} & 76.18 & 58.58 \textcolor{blue!80!black}{\scriptsize $\downarrow_{2.36}$} & 78.01 \textcolor{blue!80!black}{\scriptsize $\downarrow_{1.17}$} \\
     & \textbf{STFT} & \underline{72.24} & \underline{59.53 \textcolor{red}{\scriptsize $\uparrow_{0.17}$}} & 72.46 \textcolor{blue!80!black}{\scriptsize $\downarrow_{0.29}$} & \underline{71.99} & \underline{59.53 \textcolor{red}{\scriptsize $\uparrow_{0.21}$}} & \underline{72.54 \textcolor{red}{\scriptsize $\uparrow_{1.19}$}} & 78.47 & 60.37 \textcolor{blue!80!black}{\scriptsize $\downarrow_{0.57}$} & 80.01 \textcolor{red}{\scriptsize $\uparrow_{0.83}$} \\
     & \textbf{DFL} & 70.65 & 58.14 \textcolor{blue!80!black}{\scriptsize $\downarrow_{1.22}$} & 71.71 \textcolor{blue!80!black}{\scriptsize $\downarrow_{1.04}$} & 70.02 & 58.26 \textcolor{blue!80!black}{\scriptsize $\downarrow_{1.06}$} & 71.35 \textcolor{blue!80!black}{\scriptsize $\downarrow_{0.00}$} & \underline{78.93} & 60.51 \textcolor{blue!80!black}{\scriptsize $\downarrow_{0.43}$} & \underline{80.73 \textcolor{red}{\scriptsize $\uparrow_{1.55}$}} \\
     & \textbf{SEEN} & 68.21 & 56.06 \textcolor{blue!80!black}{\scriptsize $\downarrow_{3.30}$} & \underline{74.91 \textcolor{red}{\scriptsize $\uparrow_{2.16}$}} & 71.82 & 59.34 \textcolor{red}{\scriptsize $\uparrow_{0.02}$} & 71.60 \textcolor{red}{\scriptsize $\uparrow_{0.25}$} & 78.58 & \underline{61.03 \textcolor{red}{\scriptsize $\uparrow_{0.09}$}} & 78.79 \textcolor{blue!80!black}{\scriptsize $\downarrow_{0.39}$} \\
    \rowcolor{highlightblue} & \textbf{EchoDistill} & \textbf{73.65} & \textbf{59.57 \textcolor{red}{\scriptsize $\uparrow_{0.21}$}} & \textbf{80.32 \textcolor{red}{\scriptsize $\uparrow_{7.57}$}} & \textbf{75.36} & \textbf{62.96 \textcolor{red}{\scriptsize $\uparrow_{3.64}$}} & \textbf{76.14 \textcolor{red}{\scriptsize $\uparrow_{4.79}$}} & \textbf{79.07} & \textbf{61.79 \textcolor{red}{\scriptsize $\uparrow_{0.85}$}} & \textbf{81.37 \textcolor{red}{\scriptsize $\uparrow_{2.19}$}} \\
    \bottomrule
    \end{tabular}%
    }
\end{table*}

\vspace{-1.0em}

\subsection{Experimental Setup}
\paragraph{Models and data.}
We use a clean-teacher/noisy-student setting. The teacher and student share the same backbone; the teacher receives clean audio and is frozen, while the student receives noisy audio and is updated by EchoDistill. Main experiments cover \textit{Qwen2.5-Omni} \cite{Qwen2.5-Omni-Report}, \textit{MiniCPM-o-2.6} \cite{yao2024minicpm}, and \textit{Step-Audio-2-mini} \cite{wu2025step}; ablations are conducted on \textit{Qwen2.5-Omni}. Training and validation use paired clean/noisy audio from MMAR \cite{ma2025mmar}, with 14,397 training samples and 1,704 validation samples. The data cover 10 noise types \cite{hu2010tandem} and 7 SNR levels $\{-10,-5,0,5,10,20,30\}$\cite{zhang2026see}. Main evaluation is conducted on noisy MMAU \citep{sakshi2025mmau} at $\mathrm{SNR}=-10$.

\textbf{Baseline Methods and Evaluation Metrics.}
We compare EchoDistill with four representative noise-robustness methods, including the frequency-domain processing method \texttt{STFT}, the wavelet-transform method \texttt{WT}, the deep-learning-based denoising method \texttt{DFL} \cite{purwins2019deep}, and the representation-level noise suppression method \texttt{SEE} ~\cite{zhang2026see}. Specifically, \texttt{STFT} and \texttt{WT} represent traditional signal-processing-based front-end denoising methods, \texttt{DFL} represents a learning-based speech enhancement method, and \texttt{SEE} directly suppresses noise-related components in the internal representation space of LALMs. For fair comparison, all methods follow the same answer extraction protocol. We report three metrics: \texttt{Acc} \cite{hendrycks2020measuring}, \texttt{Noisy}\cite{zhang2026see}, and \texttt{GSR} \cite{benesty2009pearson}. Further implementation and evaluation details are provided in Appendix~\ref{app:protocol}.

\subsection{Effectiveness and Attribution (RQ1)}
\label{sec:rq1}

\textbf{Obs.\ding{182}: \ourmethod{} improves aggregate robustness across backbones
and domains, although the gains are heterogeneous.}
As shown in Table~\ref{tab:EchoDistill_Full_Metrics}, \ourmethod{} achieves
an average noisy-input accuracy of \textbf{61.44\%} across the nine
backbone--domain settings, outperforming the strongest signal- or
representation-level baseline by \textbf{1.63} percentage points (pp). Compared with
the strongest corresponding baseline on each backbone, the average gains are
\textbf{0.04} pp on \textit{Step-Audio-2-mini}, \textbf{3.43} pp on \textit{Qwen2.5-Omni}, and \textbf{0.76} pp on
\textit{MiniCPM-o-2.6}. The improvement is most pronounced for \textit{Qwen2.5-Omni} in the
Sound domain, whereas signal-processing baselines remain competitive in
Music. Moreover, \ourmethod{} improves the auxiliary \texttt{Noisy} and
\texttt{GSR} metrics by \textbf{1.80} and \textbf{4.18} pp. These results
demonstrate an aggregate improvement under severe additive noise, rather than
a uniform advantage in every backbone--domain setting. 

\begin{table}[t]
    \centering
    \footnotesize
    \setlength{\tabcolsep}{3.2pt}
    \renewcommand{\arraystretch}{1.10}
    \caption{
    Controlled noisy-data baselines and clean-capability retention on
    3,600 paired \textit{Qwen2.5-Omni} examples at
    $\mathrm{SNR}=-10\,\mathrm{dB}$.
    All values are percentages except
    $\Delta_{\mathrm{clean}}$, which is measured in percentage points.
    Invalid or unparsable outputs are counted as incorrect.
    Gap closed denotes the fraction of the Initial model's clean-to-noisy
    accuracy gap recovered by each method.
    }
    \label{tab:rq1_sft_clean_retention}

    \begin{tabularx}{\columnwidth}{
        @{}
        >{\raggedright\arraybackslash}X
        *{4}{>{\centering\arraybackslash}X}
        @{}
    }
        \toprule
        \textbf{Method}
        & \shortstack{\textbf{Noisy-Acc.}}
        & \shortstack{\textbf{Clean-Acc.}}
        & $\boldsymbol{\Delta}_{\mathrm{\mathbf{clean}}}$
        & \shortstack{\textbf{Gap-closed}} \\
        \midrule

        Initial
        & 59.33
        & 76.56
        & --
        & 0.00 \\

        Noisy-SFT
        & 60.31
        & 76.56
        & $+0.00$
        & 5.65 \\

        NoiseAug-SFT
        & 60.00
        & 76.56
        & $+0.00$
        & 3.87 \\

        \rowcolor{highlightblue}
        \textbf{\ourmethod{}}
        & \textbf{62.94}
        & \textbf{77.56}
        & $\mathbf{+1.00}$
        & \textbf{20.97} \\

        \bottomrule
    \end{tabularx}
    \vspace{-0.5\baselineskip}
\end{table}

\begin{table*}[t]
    \centering
    \small
    \setlength{\tabcolsep}{5.0pt}
    \renewcommand{\arraystretch}{1.08}
    \caption{Single-run component analysis on \textit{Qwen2.5-Omni} at
    $\mathrm{SNR}=-10\,\mathrm{dB}$. Current/old denotes the conventional
    GRPO policy ratio, while noisy/clean denotes EchoDistill's student--teacher
    ratio. \texttt{Acc} is the primary metric; \texttt{Noisy} and
    \texttt{GSR} are auxiliary agreement diagnostics.}
    \label{tab:rq1_ablation}
    \resizebox{\textwidth}{!}{%
    \begin{tabular}{lccccrrr}
        \toprule
        Variant & Policy ratio & Teacher objective & Reward bonus & Gate & Acc & Noisy & GSR \\
        \midrule
        GRPO-only & Current/old & -- & -- & -- & 59.94 & 71.47 & 71.61 \\
        Distill-only & -- & Token & -- & -- & 61.15 & 73.27 & 74.19 \\
        Standard GRPO+KL & Current/old & KL & -- & -- & 61.50 & 70.44 & 74.31 \\
        EchoDistill-StdRatio & Current/old & Token & \checkmark & \checkmark &
        60.97 & 73.58 & 74.22 \\
        EchoDistill-Ungated & Noisy/clean & Token & \checkmark & -- &
        61.03 & 73.36 & 74.17 \\
        \rowcolor{highlightblue}
        \textbf{EchoDistill} & Noisy/clean & Token & \checkmark & \checkmark &
        \textbf{62.94} & \textbf{75.36} & \textbf{76.14} \\
        \bottomrule
    \end{tabular}%
    }
    \vspace{-0.5\baselineskip}
\end{table*}
\textbf{Obs.\ding{183}: Single noisy-data SFT does not account for the improvement of \ourmethod{}.}
As shown in Table~\ref{tab:rq1_sft_clean_retention}, \ourmethod{} achieves
a noisy-input accuracy of \textbf{62.94\%}, compared with
\textbf{60.31\%} for Noisy-SFT and \textbf{60.00\%} for NoiseAug-SFT.
The corresponding paired advantages are
\textbf{2.64} pp and
\textbf{2.94} pp, respectively.
Because the controlled baselines are also trained with noisy audio, these
results indicate that exposure to corrupted inputs alone is insufficient to
reproduce the improvement. The additional gain is consistent with the benefit
of aligning noisy-input trajectories with clean-conditioned token
distributions.

\noindent
\textbf{Obs.\ding{184}: The robustness gain does not incur a clean-audio alignment
tax on the evaluated backbone.}
While noisy-input accuracy increases from \textbf{59.33\%} to
\textbf{62.94\%}, clean-audio accuracy also increases from
\textbf{76.56\%} to \textbf{77.56\%}, corresponding to a
\textbf{1.00} pp gain.
\ourmethod{} closes \textbf{20.97\%} of the Initial model's
clean-to-noisy accuracy gap without reducing its clean-audio performance.

Taken together, these controlled comparisons show that the improvement of
\ourmethod{} is neither explained by noisy-data exposure alone nor obtained
at the expense of clean-audio capability. We next examine which optimization
and distillation components contribute to this advantage.

\textbf{Obs.\ding{185}: The full formulation achieves the strongest point estimates
among the controlled optimization variants.}
As shown in Table~\ref{tab:rq1_ablation}, \ourmethod{} achieves an average
accuracy of \textbf{62.94\%}, compared with \textbf{59.94\%} for GRPO-only, \textbf{61.15\%} for
Distill-only, and \textbf{61.50\%} for Standard GRPO+KL. It therefore exceeds the
standard GRPO formulation with clean-teacher KL regularization by \textbf{1.46} pp.
Replacing the noisy/clean student--teacher policy ratio with the conventional
current/old ratio reduces accuracy to \textbf{60.97\%}, while removing the
task-positive gate yields \textbf{61.03\%}.

\begin{wrapfigure}{r}{0.56\textwidth}
    \vspace{-1.4em}
    \centering

    \includegraphics[
        width=0.92\linewidth
    ]{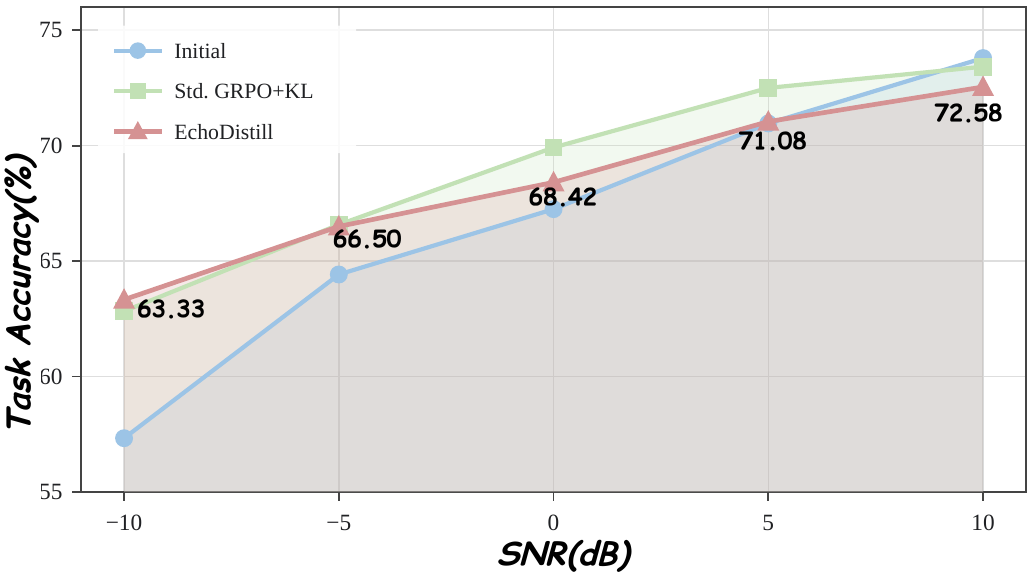}

    \vspace{-0.8em}
    \caption{%
        Task accuracy across SNR levels on \textit{Qwen2.5-Omni}. All methods use the same noisy waveforms at each SNR
        ($N=1{,}200$); annotations report \ourmethod{} experiment results.
    }
    \label{fig:rq2_snr}

    \vspace{1.0em}

    \captionof{table}{%
        Cross-benchmark accuracy (\%) at $-10\,\mathrm{dB}$.
        $\Delta_b$ denotes EchoDistill minus baseline $b$ in percentage
        points; paired 95\% bootstrap CIs are reported below.
    }
    \label{tab:rq2_external_boundary}

    \vspace{0.2em}

    {\fontsize{9.2}{10.5}\selectfont
    \renewcommand{\arraystretch}{1.18}
    \setlength{\tabcolsep}{2.5pt}
    \newcommand{\ecci}[1]{%
        {\fontsize{8.2}{8.8}\selectfont $#1$}%
    }

    \begin{tabularx}{\linewidth}{
        @{}
        >{\raggedright\arraybackslash}p{0.25\linewidth}
        *{3}{>{\centering\arraybackslash}X}
        @{}
    }
        \toprule
        \textbf{Method}
        & \textbf{AIR-Bench}
        & \textbf{DREAM-TTS}
        & \textbf{MuChoMusic} \\
        \midrule

        Init.
        & 40.73
        & 62.00
        & 55.25 \\

        GRPO+KL
        & 41.74
        & 60.75
        & 55.25 \\

        \rowcolor{highlightblue}
        \textbf{EchoDistill}
        & \textbf{42.65}
        & \textbf{66.50}
        & \textbf{57.50} \\

        \midrule

        $\Delta_{\mathrm{Init.}}$
        & $+1.92$
        & $+4.50$
        & $+2.25$ \\
        &
        \ecci{[-0.25,\,4.09]}
        & \ecci{[0.75,\,8.50]}
        & \ecci{[-1.25,\,5.75]} \\

        \addlinespace[2pt]

        $\Delta_{\mathrm{GRPO+KL}}$
        & $+0.92$
        & $+5.75$
        & $+2.25$ \\
        &
        \ecci{[-1.09,\,3.01]}
        & \ecci{[2.50,\,9.25]}
        & \ecci{[-0.75,\,5.25]} \\

        \bottomrule
    \end{tabularx}
    }

    \vspace{-1.5em}
\end{wrapfigure}

\subsection{Generalization and Limits (RQ2)}
\label{sec:rq2}

\textbf{Obs.\ding{186}: Gains concentrate under severe noise and transfer weakly to
unseen additive noise.}
In Figure~\ref{fig:rq2_snr}, EchoDistill improves over Initial by
6.00 pp at $-10\,\mathrm{dB}$, but the gain falls to \textbf{2.08, 1.17, and 0.08} pp
at $-5$, 0, and $5\,\mathrm{dB}$, respectively, and becomes \textbf{-1.25} pp at
$10\,\mathrm{dB}$. None of its paired differences from Standard GRPO+KL has
a \textbf{95\%} confidence interval excluding zero. Thus, the benefit over Initial is
concentrated at low SNR rather than uniform across noise levels.

The held-out categories show the same pattern
(Appendix Figure~\ref{fig:rq2_heldout_appendix}): the gain is \textbf{+2.00} pp at
$-10\,\mathrm{dB}$, but only \textbf{+0.81} pp when pooling all three SNRs. This provides
suggestive evidence of transfer under severe unseen additive noise, not
uniform out-of-distribution generalization.
\begin{table}[t]
    \centering
    \footnotesize
    \setlength{\tabcolsep}{2.8pt}
    \renewcommand{\arraystretch}{1.12}
    \caption{Accuracy (\%) under audio interventions on the same 1,200
    \textit{Qwen2.5-Omni} questions at $-10\,\mathrm{dB}$. The final three columns
    report matched-minus-control differences in percentage points; bracketed
    values on the second line are paired 95\% bootstrap confidence intervals.}
    \label{tab:rq3_acoustic_controls}

    \begin{tabularx}{\linewidth}{
        @{}
        l
        *{7}{>{\centering\arraybackslash}X}
        @{}
    }
        \toprule
        \textbf{Model}
        & \textbf{Matched}
        & \textbf{Random}
        & \textbf{Shuffled}
        & \textbf{Silence}
        & \shortstack{\textbf{Matched}\\$\boldsymbol{-}$\textbf{Random}}
        & \shortstack{\textbf{Matched}\\$\boldsymbol{-}$\textbf{Shuffled}}
        & \shortstack{\textbf{Matched}\\$\boldsymbol{-}$\textbf{Silence}} \\
        \midrule

        Initial
        & 57.33
        & 52.00
        & 54.00
        & 55.00
        & \shortstack{+5.33\\\mbox{[3.17, 7.58]}}
        & \shortstack{+3.33\\\mbox{[1.58, 5.08]}}
        & \shortstack{+2.33\\\mbox{[$-$0.42, 5.08]}} \\

        \rowcolor{highlightblue}
        \textbf{EchoDistill}
        & \textbf{63.33}
        & 56.92
        & 60.25
        & 59.33
        & \shortstack{\textbf{+6.42}\\\mbox{[4.25, 8.58]}}
        & \shortstack{\textbf{+3.08}\\\mbox{[1.42, 4.75]}}
        & \shortstack{\textbf{+4.00}\\\mbox{[1.50, 6.50]}} \\

        \bottomrule
    \end{tabularx}
\end{table}

\textbf{Obs.\ding{187}: Transfer is benchmark-dependent and does not reliably extend
to the tested non-additive distortions.}
Table~\ref{tab:rq2_external_boundary} shows positive point differences over
both Initial and Standard GRPO+KL on AIR-Bench Foundation
\citep{yang2024air}, DREAM-TTS, and MuChoMusic
\citep{wang2025audiobench}. Only DREAM-TTS has intervals excluding zero:
\textbf{+4.50} pp over Initial and \textbf{+5.75} pp over Standard GRPO+KL. Sample-weighted
clean accuracy is nearly unchanged (\textbf{72.57\% vs. 72.52\%}), so the evidence
supports benchmark-dependent rather than uniform transfer.
The non-additive stress test further marks the method's boundary. Across 15
bandwidth, filtering, clipping, codec, and reverberation conditions,
EchoDistill changes macro accuracy by \textbf{-0.47} pp,
with lower point estimates in 10 conditions. 

\subsection{Dependence on Matched Acoustic Evidence (RQ3)}
\label{sec:rq3}

\textbf{Obs.\ding{188}: Predictions depend measurably on question-matched acoustic
evidence.}
As shown in Table~\ref{tab:rq3_acoustic_controls}, replacing matched audio
with Random, Shuffled, or Silence reduces \ourmethod{} accuracy by \textbf{6.42, 3.08,
and 4.00} points, and all paired intervals exclude zero. Because
these effects use ground-truth correctness rather than GSR or clean--noisy agreement, an audio-independent explanation is insufficient. However,
Initial also benefits from matched audio: its corresponding gaps are \textbf{5.33,
3.33, and 2.33} points. The increase in the matched-minus-control gap after
training is therefore only \textbf{+1.09, $-$0.25, and +1.67} points.
The controls establish that matched evidence contributes to \ourmethod{}
predictions, but not that stronger acoustic dependence explains its entire
post-training gain; mismatched audio may itself be harmful.

\begin{table}[t]
    \centering
    \caption{Complementary acoustic-evidence checks in percentage points
    with 95\% bootstrap CIs.}
    \label{tab:rq3_auxiliary_controls}

    {\fontsize{10}{9}\selectfont
    \setlength{\tabcolsep}{3pt}
    \renewcommand{\arraystretch}{1.08}

    \begin{tabular*}{\linewidth}{
        @{\extracolsep{\fill}}
        l
        c
        l
        c
        >{\columncolor{highlightblue}\centering\arraybackslash}r
        r
        @{}
    }
        \toprule
        \textbf{Evaluation}
        & $N$
        & \textbf{Metric / setting}
        & \textbf{Reference}
        & \textbf{Effect}
        & \textbf{95\% CI} \\
        \midrule

        \multirow{2}{*}{AIR-Bench Chat}
        & \multirow{2}{*}{300}
        & Matched accuracy
        & Random
        & \textbf{$+12.00$}
        & $[5.67,\,18.67]$ \\

        &
        & Matched accuracy
        & Silence
        & \textbf{$+13.67$}
        & $[7.00,\,20.67]$ \\

        \cmidrule(lr){1-6}

        \multirow{4}{*}{ECAC}
        & \multirow{4}{*}{470}
        & Critical accuracy
        & GRPO+KL
        & \textbf{$+4.26$}
        & $[1.91,\,6.60]$ \\

        &
        & Drop reduction
        & GRPO+KL
        & \textbf{$+2.98$}
        & $[0.21,\,5.74]$ \\

        &
        & Evidence preservation index
        & GRPO+KL
        & \textbf{$+3.19$}
        & $[0.85,\,5.53]$ \\

        &
        & Critical accuracy
        & Initial
        & $-0.21$
        & $[-2.55,\,2.13]$ \\

        \cmidrule(lr){1-6}

        \multirow{3}{*}{Resampling}
        & \multirow{3}{*}{$600{\times}2$}
        & Realization 1
        & GRPO+KL
        & $+1.00$
        & $[-2.00,\,4.00]$ \\

        &
        & Realization 2
        & GRPO+KL
        & $+1.33$
        & $[-1.67,\,4.17]$ \\

        &
        & Pooled
        & GRPO+KL
        & $+1.17$
        & $[-1.33,\,3.67]$ \\

        \bottomrule
    \end{tabular*}
    }

    \vspace{-0.5\baselineskip}
\end{table}

\begin{wraptable}{r}{0.50\textwidth}
\vspace{-1.2em}
\centering
\caption{Teacher-error and task-positive-gate audit at
$-10\,\mathrm{dB}$ on 3,596 paired examples. WT imitation denotes
agreement with an incorrect teacher; WT recovery denotes accuracy when
the teacher is wrong. Blocked denotes verified-wrong teacher signals
excluded from the reward.}
\label{tab:rq3_teacher_gate}
\vspace{0.3em}

{\fontsize{9.5}{10}\selectfont
\setlength{\tabcolsep}{1.8pt}
\renewcommand{\arraystretch}{1.22}

\begin{tabularx}{\linewidth}{
    @{}
    >{\raggedright\arraybackslash}p{0.25\linewidth}
    *{4}{>{\centering\arraybackslash}X}
    @{}
}
\toprule
\textbf{Method} &
\shortstack{\textbf{Acc.}\\(\%)} &
\shortstack{\textbf{WT}\\\textbf{imit.}} &
\shortstack{\textbf{WT}\\\textbf{recover}} &
\shortstack{\textbf{Wrong}\\\textbf{blocked}} \\
\midrule

Initial
& 59.40
& 33.87
& 39.80
& -- \\

GRPO+KL
& 61.54
& 32.68
& \textbf{43.30}
& 98.01 \\

\rowcolor{highlightblue}
\textbf{EchoDistill}
& \textbf{62.99}
& 34.15
& 42.11
& 97.79 \\

\bottomrule
\end{tabularx}
}

\vspace{-1.2em}
\end{wraptable}

\textbf{Obs.\ding{189}: The auxiliary controls support a qualified grounding claim
and expose its boundary.}
Table~\ref{tab:rq3_auxiliary_controls} shows that, on open-ended AIR-Bench
Chat, matched audio improves judged correctness over Random and Silence by
\textbf{12.00 and 13.67} points, reducing the likelihood that the main effect is solely an answer-position shortcut. Under ECAC, EchoDistill exceeds Standard GRPO+KL
by \textbf{4.26} points on the Critical condition, reduces the clean-to-Critical drop
by \textbf{2.98} points, and improves EPI by \textbf{3.19} points. Its Critical accuracy is, however, statistically indistinguishable from Initial (\textbf{$-0.21$} points), so ECAC supports robustness over the controlled optimizer rather than universally stronger grounding. Table~\ref{tab:rq3_teacher_gate} further shows that, when the clean teacher is wrong, \ourmethod{} recovers the correct answer more often
than it imitates the teacher (\textbf{42.11\% vs. 34.15\%}), while the task-positive
gate blocks \textbf{97.79\%} of verified-wrong teacher signals. Finally, both new noise realizations retain positive point differences, but the pooled interval for
the \textbf{+1.17-point} gain includes zero. Thus, matched evidence and selective
teacher guidance contribute to the learned behavior. Full transition counts, ECAC results, option-rotation results, and
teacher-confidence breakdowns are reported in the Appendix~\ref{app:rq3}.

\vspace{-0.6em}
\section{Conclusion }
\vspace{-0.6em}

In this work, we propose \ourmethod, a noisy-to-clean self-distillation framework for improving LALM robustness under severe acoustic noise. By combining a frozen clean-audio teacher with noisy student rollouts, token-level alignment, and audio-aware reward shaping, \ourmethod improves noisy-input accuracy and generation stability without degrading clean-audio capability or adding inference-time costs. Experiments across three LALMs, together with controlled baselines and audio interventions, demonstrate the effectiveness of noisy-to-clean semantic alignment and confirm that matched acoustic evidence contributes to model predictions.

\newpage 
\section*{AI Use Statement}

We used AI tools to assist with language polishing and manuscript editing.
The authors remain fully responsible for the technical content, experimental
results, analyses, citations, and claims presented in this paper.

\section*{Ethics Statement}

\ourmethod{} uses existing audio datasets and pretrained models to study the
noise robustness of large audio language models. Model predictions and
clean-conditioned teacher signals may contain errors or reflect biases in the
underlying data and models. Applications involving personal audio should
respect data licenses, consent, and privacy, and should include safeguards
against unsupported or sensitive inferences about individuals.

\section*{Reproducibility Statement}

The model roles, datasets, baselines, and evaluation settings are described in
Section~\ref{experiments}. The training objective and inference procedure
are detailed in Section~\ref{sec:method}. Additional information on data
construction, metric definitions, statistical analysis, controlled audio
interventions, and supplementary results is provided in
Appendices~\ref{app:protocol}--\ref{app:rq3}.

\bibliography{iclr2027_conference}
\bibliographystyle{iclr2027_conference}
\newpage
\appendix
\section{Appendix}

\subsection{Extended Related Work}
\label{app:related_work}

\paragraph{Group Relative Policy Optimization.}
Group Relative Policy Optimization (GRPO)
\citep{shao2024deepseekmath, song2026expanding} samples a group of responses
$\{y^{(k)}\}_{k=1}^{K}$ and normalizes their rewards within the group:
\begin{equation}
A^{(k)}
=
\frac{r^{(k)}-\mu}{\sigma+\varepsilon},
\end{equation}
where $\mu$ and $\sigma$ are the group reward mean and standard deviation.
For
\begin{equation}
\rho_t^{(k)}
=
\frac{
\pi_{\theta}(y_t^{(k)}\mid x,y_{<t}^{(k)})
}{
\pi_{\theta_{\mathrm{old}}}(y_t^{(k)}\mid x,y_{<t}^{(k)})
},
\end{equation}
the clipped objective is
\begin{align}
\mathcal{L}_{\mathrm{GRPO}}
=
-\frac{1}{K}\sum_{k=1}^{K}
\frac{1}{|y^{(k)}|}
\sum_{t=1}^{|y^{(k)}|}
\min\Big(
&\rho_t^{(k)}A^{(k)}, \operatorname{clip}
(\rho_t^{(k)},1-\epsilon,1+\epsilon)A^{(k)}
\Big).
\end{align}
GRPO avoids training a separate value model, but its sequence-level advantage
is shared across tokens and therefore does not identify which token decisions
are altered by acoustic corruption.

\paragraph{Robust speech processing.}
Robust speech systems have long addressed acoustic mismatch through
multi-condition training, waveform augmentation, spectral masking, speech
enhancement, and source separation
\citep{li2015robust,ko2015audio,park2019specaugment,
pascual2017segan,wang2018supervised}.
These methods are effective when improved signal fidelity or recognition
accuracy is closely coupled with the downstream objective. However, LALMs
must map acoustic observations into open-ended semantic responses, so
recovering a perceptually cleaner signal does not necessarily preserve the
specific evidence required by a downstream question. Moreover, aggressive
enhancement may suppress weak but discriminative acoustic cues when those
cues overlap with interference \citep{purwins2019deep}.

\paragraph{Noise robustness for LALMs.}
Acoustic corruption can change the high-level predictions of LALMs even when
their generated language remains fluent
\citep{li2025isa,hou2025evaluating,wang2025audiobench}.
Recent methods use text-queried separation, plug-and-play audio enhancement,
and representation-level suppression to reduce this sensitivity
\citep{yin2024leveraging,chen2025noise,yin2026focus,zhang2026see}.
These approaches mainly intervene in the waveform, encoder features, or
internal noise representations. Such intervention can be complementary to
\ourmethod{}, but it does not explicitly optimize the semantic decisions
made along noisy-input generation trajectories. \ourmethod{} instead retains
the original input and model architecture at inference time and uses paired
clean audio only to construct privileged post-training supervision.

\paragraph{Knowledge distillation and on-policy learning.}
Classical knowledge distillation matches softened predictions or internal
representations from a teacher
\citep{hinton2015distilling}.
For autoregressive models, offline teacher forcing can expose the student to
prefixes that differ from those encountered during its own generation,
creating a distribution mismatch related to exposure bias
\citep{ross2011reduction}.
On-policy distillation addresses this issue by evaluating teacher guidance on
student-generated trajectories
\citep{agarwal2024policy}.
Recent studies further explore self-distillation between checkpoints or
different states of the same model
\citep{zhao2026self,hubotter2026reinforcement,
shenfeld2026self,ko2026scaling}.
\ourmethod{} combines on-policy guidance with asymmetric acoustic information:
the clean and noisy branches evaluate the same student-generated prefix but
receive different versions of the audio. This paired comparison estimates
where corruption changes the model's next-token preference. Task-positive
gating further prevents agreement with an incorrect clean-conditioned
prediction from being rewarded indiscriminately.

\subsection{Experimental Protocol and Metrics}
\label{app:protocol}

\paragraph{Models and paired data.}
We evaluate Qwen2.5-Omni, MiniCPM-o-2.6, and Step-Audio2
\citep{Qwen2.5-Omni-Report,yao2024minicpm,wu2025step}. For each backbone, the teacher
and student are initialized from the same checkpoint, after which the teacher
is frozen. Training and validation contain 14,397 and 1,704 paired clean/noisy
examples from MMAR \citep{ma2025mmar}. Training corruptions comprise ten
additive-noise categories and seven SNR levels,
$\{-10,-5,0,5,10,20,30\}\,\mathrm{dB}$. The primary evaluation contains 3,600
MMAU examples \citep{sakshi2025mmau}, equally divided among Music, Sound,
and Speech. Unless stated otherwise, results use additive noise at
$-10\,\mathrm{dB}$.

\paragraph{Primary and auxiliary metrics.}
Noisy-input task accuracy is the primary metric. For a fixed evaluation set of
size $N$, invalid or unparsable answers are counted as incorrect:
\begin{equation}
    \mathrm{Acc}_{d}^{\mathrm{n}}
    = \frac{1}{N}\sum_{i=1}^{N}
      \mathbb{I}\!\left[\hat y_{i}^{d,\mathrm{n}}=y_i^\star\right].
    \label{eq:app_noisy_accuracy}
\end{equation}
Clean-audio accuracy $\mathrm{Acc}_{d}^{\mathrm{c}}$ is defined analogously.
Let $d=0$ denote the initial model. We report
\begin{align}
    \Delta_{\mathrm{clean}}(d)
    &= \mathrm{Acc}_{d}^{\mathrm{c}}-\mathrm{Acc}_{0}^{\mathrm{c}}, \\
    \mathrm{GapClosure}(d)
    &= \frac{\mathrm{Acc}_{d}^{\mathrm{n}}-\mathrm{Acc}_{0}^{\mathrm{n}}}
            {\mathrm{Acc}_{0}^{\mathrm{c}}-\mathrm{Acc}_{0}^{\mathrm{n}}}
       \times 100\%.
    \label{eq:app_gap_closure}
\end{align}

For comparison with prior work, we retain \texttt{Noisy} and \texttt{GSR} as
auxiliary agreement diagnostics. Let $\hat y_i^{0,\mathrm{c}}$ be the initial
model's clean-audio prediction and let $\mathcal{V}_{\mathrm{Noisy}}$ and
$\mathcal{V}_{\mathrm{GSR}}$ denote the corresponding valid paired-output
sets. We compute
\begin{align}
    \mathrm{Noisy}_{d}
    &= \frac{1}{|\mathcal{V}_{\mathrm{Noisy}}|}
       \sum_{i\in\mathcal{V}_{\mathrm{Noisy}}}
       \mathbb{I}\!\left[\hat y_i^{d,\mathrm{n}}
                         =\hat y_i^{0,\mathrm{c}}\right], \\
    \mathrm{GSR}_{d}
    &= \frac{1}{|\mathcal{V}_{\mathrm{GSR}}|}
       \sum_{i\in\mathcal{V}_{\mathrm{GSR}}}
       \mathbb{I}\!\left[\hat y_i^{d,\mathrm{n}}
                         =\hat y_i^{d,\mathrm{c}}\right].
    \label{eq:app_agreement_metrics}
\end{align}
These diagnostics quantify agreement with clean-conditioned behavior. They
are not correctness or grounding metrics: predictions may agree while both
are wrong, and an input-independent model may also obtain high agreement.

\paragraph{Statistical analysis.}
Unless otherwise stated, intervals are paired 95\% bootstrap confidence
intervals based on 5,000 resamples with seed 42. When several corruptions are
derived from the same clean source, resampling is clustered by source.
Corrected/broken transitions are additionally evaluated with the exact
McNemar test. The independent noise-resampling experiment uses 10,000 paired
cluster-bootstrap resamples. These intervals capture evaluation-sample
uncertainty conditional on one trained checkpoint and the evaluated
waveforms; they do not capture post-training seed variability.

\paragraph{Training-instance format.}
Each record contains a paired waveform, the task prompt, candidate choices,
the target, and corruption metadata. Student candidates, teacher scores, and
rewards are produced online rather than stored as fixed targets.

\paragraph{Cross-table denominators.}
Table~1 reports 62.96\% because its Average row is the macro-average
of the three domain-level accuracies. Tables~2, 3, and~8 instead report
pooled accuracy on the same fixed set of 3,600 examples, yielding
62.94\%, with invalid or unparsable outputs counted as incorrect.
Table~13 reports 62.99\% because the teacher-reliability audit is
restricted to 3,596 eligible examples. These small differences arise
from aggregation and evaluation eligibility rather than different
model checkpoints.

\subsection{Limitations}
\ourmethod{} relies on paired clean--noisy audio and an imperfect clean-conditioned teacher, while noisy rollouts and token-level scoring increase post-training cost. Its gains are strongest under severe additive noise and do not reliably extend to the tested non-additive distortions. Moreover, several analyses use a single trained checkpoint, so future work should examine seed variability, broader acoustic conditions, and more natural open-ended audio interactions.

\subsection{Illustrative Training Record}

To qualitatively examine whether the model grounds its predictions in the
acoustic input rather than relying primarily on linguistic priors, we present
three representative case studies. Case Study 1 illustrates the structure of
a noisy-audio training record. Case Studies 2 and 3 compare the responses of
the initial model, standard GRPO+KL, and \ourmethod{} under the same noisy input.
We additionally replace the relevant waveform with random audio or remove it
entirely. These controlled variants reveal whether a correct response is
supported by the acoustic evidence: \ourmethod{} recovers the reference answer
from the noisy waveform, whereas its predictions change when the audio is
irrelevant or unavailable.

The examples were manually selected for interpretability and should not be
viewed as an unbiased sample of model successes or failures.
\newpage
\begin{figure*}[t!]
    \centering
    \begin{tcolorbox}[
        colback=orange!5!white,
        colframe=cyan!75!black,
        title={Case Study 1},
        fonttitle=\bfseries,
        sharp corners,
        boxrule=0.8pt,
        left=2mm,
        right=2mm,
        top=2mm,
        bottom=2mm
    ]
        \textbf{Example ID: 19452}\\
        \vspace{0.4em}

        \textit{\textbf{Prompt:}}
        What is producing the sound in the audio?\\

        \textit{\textbf{Choices:}}
        Airplane, Motorcycle, Train, Sports car\\

        \textit{\textbf{Acoustic condition:}}
        Water noise at an SNR of 30\,dB.\\

        \textit{\textbf{Target:}}
        \textbf{Airplane}\\[0.5em]

        \textbf{Training Record (Raw Input Format):}

        \begin{tcolorbox}[
            colback=black!5!white,
            colframe=gray!20,
            boxrule=0.5pt,
            arc=0pt,
            left=1mm,
            right=1mm,
            top=1mm,
            bottom=1mm,
            fontupper=\small\ttfamily
        ]
        \raggedright
        \{\\
        \hspace*{1em}"id": 19452,\\
        \hspace*{1em}"prompt":
        "What is producing the sound in the audio?",\\
        \hspace*{1em}"noisy\_audio\_path":
        ".../noise/water/snr\_30/...wav",\\
        \hspace*{1em}"clean\_audio\_path":
        ".../audio/...wav",\\
        \hspace*{1em}"choices":
        ["Airplane", "Motorcycle", "Train", "Sports car"],\\
        \hspace*{1em}"target": "\textbf{Airplane}",\\
        \hspace*{1em}"noise\_type": "water",\\
        \hspace*{1em}"snr": 30\\
        \}
        \end{tcolorbox}

        \vspace{0.3em}
        \textit{\textcolor{gray!80!black}{
        Paths are abbreviated for readability. The target field provides
        the supervision signal used during training.
        }}
    \end{tcolorbox}
    \vspace{-1.0em}
\end{figure*}

\clearpage

\begin{figure*}[t!]
    \centering
    \begin{tcolorbox}[
        colback=orange!5!white,
        colframe=cyan!75!black,
        title={Case Study 2 (1/2)},
        fonttitle=\bfseries,
        sharp corners,
        boxrule=0.8pt,
        left=2mm,
        right=2mm,
        top=2mm,
        bottom=2mm
    ]
        \textbf{Example ID: 1907}\\
        \vspace{0.4em}

        \textit{\textbf{Question:}}
        In which show did the male speaker mention a guest appearance was made?\\

        \textit{\textbf{Reference answer:}}
        Little Britain.\\[0.4em]

        \textbf{Answer\ding{182}: Initial (Noisy Audio)}

        \begin{tcolorbox}[
            colback=black!5!white,
            colframe=gray!20,
            boxrule=0.5pt,
            arc=0pt,
            left=1mm,
            right=1mm,
            top=1mm,
            bottom=1mm,
            fontupper=\small\ttfamily
        ]
        \raggedright
        \{ "prompt": "\hlToken{im\_start}system\\
        You are a helpful assistant.\hlToken{im\_end}\\
        \hlToken{im\_start}user\\
        \hlToken{audio\_bos}\hlToken{AUDIO}\hlToken{AUDIO}...
        \textit{\textcolor{gray}{(noisy audio tokens)}}
        ...\hlToken{AUDIO}\hlToken{AUDIO}\hlToken{audio\_eos}
        In which show did the male speaker mention a guest appearance was made?
        \hlToken{im\_end}\\
        \hlToken{im\_start}assistant\\
        ", "predict": "The male speaker mentioned that a guest appeared on the show
        'The Tonight Show Starring Jimmy Fallon'." \}
        \end{tcolorbox}

        \textbf{Answer\ding{183}: Standard GRPO+KL (Noisy Audio)}

        \begin{tcolorbox}[
            colback=black!5!white,
            colframe=gray!20,
            boxrule=0.5pt,
            arc=0pt,
            left=1mm,
            right=1mm,
            top=1mm,
            bottom=1mm,
            fontupper=\small\ttfamily
        ]
        \raggedright
        \{ "prompt": "\hlToken{im\_start}system\\
        You are a helpful assistant.\hlToken{im\_end}\\
        \hlToken{im\_start}user\\
        \hlToken{audio\_bos}\hlToken{AUDIO}\hlToken{AUDIO}...
        \textit{\textcolor{gray}{(noisy audio tokens)}}
        ...\hlToken{AUDIO}\hlToken{AUDIO}\hlToken{audio\_eos}
        In which show did the male speaker mention a guest appearance was made?
        \hlToken{im\_end}\\
        \hlToken{im\_start}assistant\\
        ", "predict": "The male speaker mentioned a guest appearance was made in
        the show 'The X-Files'." \}
        \end{tcolorbox}

        \textbf{Answer\ding{184}: \ourmethod{} (Noisy Audio)}

        \begin{tcolorbox}[
            colback=black!5!white,
            colframe=gray!20,
            boxrule=0.5pt,
            arc=0pt,
            left=1mm,
            right=1mm,
            top=1mm,
            bottom=1mm,
            fontupper=\small\ttfamily
        ]
        \raggedright
        \{ "prompt": "\hlToken{im\_start}system\\
        You are a helpful assistant.\hlToken{im\_end}\\
        \hlToken{im\_start}user\\
        \hlToken{audio\_bos}\hlToken{AUDIO}\hlToken{AUDIO}...
        \textit{\textcolor{gray}{(noisy audio tokens)}}
        ...\hlToken{AUDIO}\hlToken{AUDIO}\hlToken{audio\_eos}
        In which show did the male speaker mention a guest appearance was made?
        \hlToken{im\_end}\\
        \hlToken{im\_start}assistant\\
        ", "predict": "The male speaker mentioned a guest appearance was made in
        Little Britain." \}
        \end{tcolorbox}
    \end{tcolorbox}
    \vspace{-1.0em}
\end{figure*}

\clearpage

\begin{figure*}[t!]
    \centering
    \begin{tcolorbox}[
        colback=orange!5!white,
        colframe=cyan!75!black,
        title={Case Study 2 (2/2)},
        fonttitle=\bfseries,
        sharp corners,
        boxrule=0.8pt,
        left=2mm,
        right=2mm,
        top=2mm,
        bottom=2mm
    ]
        \textbf{Example ID: 1907}\\
        \vspace{0.4em}

        \textit{\textbf{Question:}}
        In which show did the male speaker mention a guest appearance was made?\\

        \textit{\textbf{Reference answer:}}
        Little Britain.\\[0.4em]

        \textbf{Answer\ding{185}: \ourmethod{} (Random Audio)}

        \begin{tcolorbox}[
            colback=black!5!white,
            colframe=gray!20,
            boxrule=0.5pt,
            arc=0pt,
            left=1mm,
            right=1mm,
            top=1mm,
            bottom=1mm,
            fontupper=\small\ttfamily
        ]
        \raggedright
        \{ "prompt": "\hlToken{im\_start}system\\
        You are a helpful assistant.\hlToken{im\_end}\\
        \hlToken{im\_start}user\\
        \hlToken{audio\_bos}\hlToken{AUDIO}\hlToken{AUDIO}...
        \textit{\textcolor{gray}{(random audio tokens)}}
        ...\hlToken{AUDIO}\hlToken{AUDIO}\hlToken{audio\_eos}
        In which show did the male speaker mention a guest appearance was made?
        \hlToken{im\_end}\\
        \hlToken{im\_start}assistant\\
        ", "predict": "The male speaker mentioned a guest appearance was made in
        the show 'The Best of the Best'." \}
        \end{tcolorbox}

        \textbf{Answer\ding{186}: \ourmethod{} (No Audio)}

        \begin{tcolorbox}[
            colback=black!5!white,
            colframe=gray!20,
            boxrule=0.5pt,
            arc=0pt,
            left=1mm,
            right=1mm,
            top=1mm,
            bottom=1mm,
            fontupper=\small\ttfamily
        ]
        \raggedright
        \{ "prompt": "\hlToken{im\_start}system\\
        You are a helpful assistant.\hlToken{im\_end}\\
        \hlToken{im\_start}user\\
        In which show did the male speaker mention a guest appearance was made?
        \hlToken{im\_end}\\
        \hlToken{im\_start}assistant\\
        ", "predict": "I'm sorry, but I cannot answer this question without more
        context. Can you please provide me with more information about the male
        speaker and the show in question?" \}
        \end{tcolorbox}
    \end{tcolorbox}
    \vspace{-1.0em}
\end{figure*}
\noindent\textbf{Takeaway.}
Under the matched noisy waveform, \ourmethod{} recovers the reference
\emph{Little Britain}, whereas replacing or removing the waveform changes the
answer. This example illustrates sensitivity to question-matched acoustic
evidence, but does not by itself establish a population-level effect.

\clearpage

\begin{figure*}[t!]
    \centering
    \begin{tcolorbox}[
        colback=orange!5!white,
        colframe=cyan!75!black,
        title={Case Study 3 (1/2)},
        fonttitle=\bfseries,
        sharp corners,
        boxrule=0.8pt,
        left=2mm,
        right=2mm,
        top=2mm,
        bottom=2mm
    ]
        \textbf{Example ID: 97}\\
        \vspace{0.4em}

        \textit{\textbf{Question:}}
        What is the speaker's final statement about the future of peace in the area?\\

        \textit{\textbf{Reference answer:}}
        The speaker's final statement is that there's never going to be peace over there.\\[0.4em]

        \textbf{Answer\ding{182}: Initial (Noisy Audio)}

        \begin{tcolorbox}[
            colback=black!5!white,
            colframe=gray!20,
            boxrule=0.5pt,
            arc=0pt,
            left=1mm,
            right=1mm,
            top=1mm,
            bottom=1mm,
            fontupper=\small\ttfamily
        ]
        \raggedright
        \{ "prompt": "\hlToken{im\_start}system\\
        You are a helpful assistant.\hlToken{im\_end}\\
        \hlToken{im\_start}user\\
        \hlToken{audio\_bos}\hlToken{AUDIO}\hlToken{AUDIO}...
        \textit{\textcolor{gray}{(noisy audio tokens)}}
        ...\hlToken{AUDIO}\hlToken{AUDIO}\hlToken{audio\_eos}
        What is the speaker's final statement about the future of peace in the area?
        \hlToken{im\_end}\\
        \hlToken{im\_start}assistant\\
        ", "predict": "The speaker believes that peace will always be a possibility
        in the area." \}
        \end{tcolorbox}

        \textbf{Answer\ding{183}: Standard GRPO+KL (Noisy Audio)}

        \begin{tcolorbox}[
            colback=black!5!white,
            colframe=gray!20,
            boxrule=0.5pt,
            arc=0pt,
            left=1mm,
            right=1mm,
            top=1mm,
            bottom=1mm,
            fontupper=\small\ttfamily
        ]
        \raggedright
        \{ "prompt": "\hlToken{im\_start}system\\
        You are a helpful assistant.\hlToken{im\_end}\\
        \hlToken{im\_start}user\\
        \hlToken{audio\_bos}\hlToken{AUDIO}\hlToken{AUDIO}...
        \textit{\textcolor{gray}{(noisy audio tokens)}}
        ...\hlToken{AUDIO}\hlToken{AUDIO}\hlToken{audio\_eos}
        What is the speaker's final statement about the future of peace in the area?
        \hlToken{im\_end}\\
        \hlToken{im\_start}assistant\\
        ", "predict": "The speaker believes that peace will always be a possibility
        in the area." \}
        \end{tcolorbox}

        \textbf{Answer\ding{184}: \ourmethod{} (Noisy Audio)}

        \begin{tcolorbox}[
            colback=black!5!white,
            colframe=gray!20,
            boxrule=0.5pt,
            arc=0pt,
            left=1mm,
            right=1mm,
            top=1mm,
            bottom=1mm,
            fontupper=\small\ttfamily
        ]
        \raggedright
        \{ "prompt": "\hlToken{im\_start}system\\
        You are a helpful assistant.\hlToken{im\_end}\\
        \hlToken{im\_start}user\\
        \hlToken{audio\_bos}\hlToken{AUDIO}\hlToken{AUDIO}...
        \textit{\textcolor{gray}{(noisy audio tokens)}}
        ...\hlToken{AUDIO}\hlToken{AUDIO}\hlToken{audio\_eos}
        What is the speaker's final statement about the future of peace in the area?
        \hlToken{im\_end}\\
        \hlToken{im\_start}assistant\\
        ", "predict": "The speaker believes that peace will never be achieved in
        the area." \}
        \end{tcolorbox}
    \end{tcolorbox}
    \vspace{-1.0em}
\end{figure*}

\clearpage

\begin{figure*}[t!]
    \centering
    \begin{tcolorbox}[
        colback=orange!5!white,
        colframe=cyan!75!black,
        title={Case Study 3 (2/2)},
        fonttitle=\bfseries,
        sharp corners,
        boxrule=0.8pt,
        left=2mm,
        right=2mm,
        top=2mm,
        bottom=2mm
    ]
        \textbf{Example ID: 97}\\
        \vspace{0.4em}

        \textit{\textbf{Question:}}
        What is the speaker's final statement about the future of peace in the area?\\

        \textit{\textbf{Reference answer:}}
        The speaker's final statement is that there's never going to be peace over there.\\[0.4em]

        \textbf{Answer\ding{185}: \ourmethod{} (Random Audio)}

        \begin{tcolorbox}[
            colback=black!5!white,
            colframe=gray!20,
            boxrule=0.5pt,
            arc=0pt,
            left=1mm,
            right=1mm,
            top=1mm,
            bottom=1mm,
            fontupper=\small\ttfamily
        ]
        \raggedright
        \{ "prompt": "\hlToken{im\_start}system\\
        You are a helpful assistant.\hlToken{im\_end}\\
        \hlToken{im\_start}user\\
        \hlToken{audio\_bos}\hlToken{AUDIO}\hlToken{AUDIO}...
        \textit{\textcolor{gray}{(random audio tokens)}}
        ...\hlToken{AUDIO}\hlToken{AUDIO}\hlToken{audio\_eos}
        What is the speaker's final statement about the future of peace in the area?
        \hlToken{im\_end}\\
        \hlToken{im\_start}assistant\\
        ", "predict": "The speaker believes that peace will be achieved in the
        future." \}
        \end{tcolorbox}

        \textbf{Answer\ding{186}: \ourmethod{} (No Audio)}

        \begin{tcolorbox}[
            colback=black!5!white,
            colframe=gray!20,
            boxrule=0.5pt,
            arc=0pt,
            left=1mm,
            right=1mm,
            top=1mm,
            bottom=1mm,
            fontupper=\small\ttfamily
        ]
        \raggedright
        \{ "prompt": "\hlToken{im\_start}system\\
        You are a helpful assistant.\hlToken{im\_end}\\
        \hlToken{im\_start}user\\
        What is the speaker's final statement about the future of peace in the area?
        \hlToken{im\_end}\\
        \hlToken{im\_start}assistant\\
        ", "predict": "The speaker's final statement about the future of peace in
        the area is that it is uncertain and depends on the actions of the people
        living there." \}
        \end{tcolorbox}
    \end{tcolorbox}
    \vspace{-1.0em}
\end{figure*}

\noindent\textbf{Takeaway.}
The matched noisy waveform allows \ourmethod{} to preserve the negative
meaning of the speaker's statement, while the Random-Audio and No-Audio
conditions reverse or weaken that conclusion. The example is illustrative;
the aggregate grounding claim is based on the paired controls in
Table~\ref{tab:app_audio_transitions}.

\newpage

\section{Additional Results for RQ1}
\label{app:rq1}

\paragraph{Fixed-denominator transition audit.}
Table~\ref{tab:app_rq1_transition} examines paired correctness transitions
on the same 3,600 Qwen2.5-Omni examples used for the clean-capability
comparison. Invalid or unparsable outputs count as incorrect for both
models. Under $-10\,\mathrm{dB}$ additive noise, \ourmethod{} corrects
\textbf{297} errors made by Initial while breaking \textbf{167} initially
correct predictions. The resulting net improvement is
\textbf{+3.61 percentage points} (95\% CI: $[2.44,4.81]$), recovering
\textbf{20.97\%} of Initial's clean-to-noisy accuracy gap. On the paired
clean inputs, accuracy increases from \textbf{76.56\%} to
\textbf{77.56\%} (95\% CI for the difference: $[0.33,1.67]$).
Thus, the noisy-input improvement is not accompanied by a clean-audio
accuracy loss on this evaluation. This clean-retention result applies to
Qwen2.5-Omni; equivalent fixed-denominator audits are not available for
the other two backbones.

\begin{table*}[t]
    \centering
    \small
    \setlength{\tabcolsep}{5.5pt}
    \renewcommand{\arraystretch}{1.08}
    \caption{Fixed-denominator Qwen2.5-Omni audit on 3,600 paired examples.
    Accuracy and gap closure are percentages; gains are percentage points.
    Corrected/Broken counts paired noisy-input transitions.}
    \label{tab:app_rq1_transition}
    \begin{tabular*}{\textwidth}{@{\extracolsep{\fill}}lrrrrll}
        \toprule
        Input & Initial Acc. & Echo Acc. & Gain & Gap Closure & 95\% CI & Corrected/Broken \\
        \midrule
        Noisy ($-10\,\mathrm{dB}$) & 59.33 & \textbf{62.94} & +3.61 & 20.97
        & [2.44, 4.81] & 297/167 \\
        Clean & 76.56 & \textbf{77.56} & +1.00 & --
        & [0.33, 1.67] & -- \\
        \bottomrule
    \end{tabular*}
\end{table*}

\section{Additional Results for RQ2}
\label{app:rq2}

\subsection{Held-Out Additive-Noise Categories}
\label{app:rq2_heldout}

Figure~\ref{fig:rq2_heldout_appendix} and
Table~\ref{tab:app_heldout_additive} test six additive-noise categories
excluded from training. The two models receive identical, reproducibly
mixed waveforms, so each comparison is paired at the source-task level.
At $-10\,\mathrm{dB}$, \ourmethod{} improves accuracy from
\textbf{64.78\%} to \textbf{66.78\%}: a \textbf{+2.00-point} difference
with a paired 95\% cluster-bootstrap CI of $[0.11,4.00]$. The exact
McNemar test yields $p=0.0535$, however, so the two uncertainty analyses
should not be summarized as an unqualified significant improvement.

The advantage becomes smaller as the corruption weakens:
\textbf{+0.78 points} at $-5\,\mathrm{dB}$ and
\textbf{$-0.33$ points} at $0\,\mathrm{dB}$. Across all three SNRs,
the pooled difference is \textbf{+0.81 points} (95\% CI:
$[-0.56,2.19]$). Although the six category-level point estimates at
$-10\,\mathrm{dB}$ are nonnegative, their individual intervals include
zero. These results suggest transfer to unseen additive noise under the
most severe tested corruption, but do not establish a uniform benefit
across noise categories or intensities. None of the six category-specific confidence intervals excludes zero;
the category-level breakdown is therefore descriptive and should not be used
to rank the relative transferability of individual noise types.

\begin{figure}[t]
    \centering
    \begin{tikzpicture}
    \begin{axis}[
        width=0.90\linewidth,
        height=5.2cm,
        xlabel={Held-out additive-noise setting},
        ylabel={\ourmethod{} $-$ Initial (pp)},
        xmin=0.5,
        xmax=4.5,
        ymin=-2.5,
        ymax=4.8,
        xtick={1,2,3,4},
        xticklabels={$-10$ dB,$-5$ dB,$0$ dB,Pooled},
        xticklabel style={
            font=\scriptsize,
            rotate=16,
            anchor=east,
            text=black!75
        },
        yticklabel style={
            font=\scriptsize,
            text=black!75
        },
        label style={
            font=\small,
            text=black!85
        },
        axis line style={
            draw=black!60,
            line width=0.7pt
        },
        tick style={
            draw=black!55
        },
        grid=major,
        grid style={
            draw=softGrid,
            line width=0.55pt
        },
    ]

    \addplot+[
        zeroGray,
        dashed,
        line width=0.9pt,
        mark=none,
        domain=0.5:4.5,
        samples=2
    ] {0};

    \addplot+[
        only marks,
        color=echoRoseDark,
        mark=square*,
        mark size=3pt,
        mark options={
            fill=echoRose,
            draw=echoRoseDark,
            line width=0.7pt
        },
        error bars/.cd,
        y dir=both,
        y explicit,
        error bar style={
            color=echoRoseDark,
            line width=1pt
        },
        error mark options={
            color=echoRoseDark,
            mark size=3pt,
            line width=0.9pt
        }
    ]
    table[
        x=x,
        y=y,
        y error minus=emin,
        y error plus=eplus
    ] {
        x  y      emin  eplus
        1  2.00   1.89  2.00
        2  0.78   1.78  1.66
        3 -0.33   1.56  1.55
        4  0.81   1.37  1.38
    };

    \end{axis}
    \end{tikzpicture}

    \caption{Transfer to six additive-noise categories excluded from training
    on Qwen2.5-Omni. Error bars indicate paired 95\% cluster-bootstrap
    confidence intervals.}
    \label{fig:rq2_heldout_appendix}
\end{figure}

\begin{table*}[t]
    \centering
    \scriptsize
    \setlength{\tabcolsep}{4.0pt}
    \renewcommand{\arraystretch}{1.08}
    \caption{Held-out additive-noise evaluation on 900 source tasks. The
    pooled row contains three SNR observations per task and clusters its
    confidence interval by source. C/B denotes corrected/broken predictions.}
    \label{tab:app_heldout_additive}

    \resizebox{\textwidth}{!}{%
    \begin{tabular}{
        l
        r
        r
        >{\columncolor{echoBlue!28}\color{echoBlueText}}r
        r
        l
        r
        r
    }
        \toprule
        Setting
        & $N$
        & Initial
        & \textbf{\ourmethod{}}
        & Gain
        & 95\% CI
        & C/B
        & McNemar $p$ \\
        \midrule

        \multicolumn{8}{l}{\textit{Aggregate by SNR}} \\

        $-10\,\mathrm{dB}$
        & 900 & 64.78 & \textbf{66.78}
        & +2.00 & [0.11, 4.00] & 48/30 & 0.0535 \\

        $-5\,\mathrm{dB}$
        & 900 & 67.89 & 68.67
        & +0.78 & [$-$1.00, 2.44] & 36/29 & 0.4570 \\

        $0\,\mathrm{dB}$
        & 900 & 71.00 & 70.67
        & $-$0.33 & [$-$1.89, 1.22] & 24/27 & 0.7798 \\

        Pooled
        & 2,700 & 67.89 & 68.70
        & +0.81 & [$-$0.56, 2.19] & 108/86 & 0.1314 \\

        \midrule
        \multicolumn{8}{l}{
            \textit{Breakdown at $-10\,\mathrm{dB}$}
        } \\

        Clock tick
        & 150 & 62.67 & 66.67
        & +4.00 & [$-$1.33, 9.33] & 11/5 & 0.2101 \\

        Crackling fire
        & 150 & 68.67 & 70.00
        & +1.33 & [$-$3.33, 6.00] & 7/5 & 0.7744 \\

        Keyboard typing
        & 150 & 66.67 & 70.00
        & +3.33 & [$-$0.67, 8.00] & 8/3 & 0.2266 \\

        Rain
        & 150 & 56.67 & 59.33
        & +2.67 & [$-$2.67, 8.00] & 10/6 & 0.4545 \\

        Sea waves
        & 150 & 62.67 & 62.67
        & 0.00 & [$-$4.67, 4.67] & 7/7 & 1.0000 \\

        Thunderstorm
        & 150 & 71.33 & 72.00
        & +0.67 & [$-$3.33, 4.67] & 5/4 & 1.0000 \\

        \bottomrule
    \end{tabular}%
    }
\end{table*}

\subsection{Non-Additive Robustness Boundary}
\label{app:rq2_nonadditive}

Table~\ref{tab:app_nonadditive} evaluates 15 non-additive conditions
formed from five distortion families and three severity levels. Unlike
the held-out additive-noise experiment, these conditions change the
signal through bandwidth restriction, filtering, clipping, codec
processing, or reverberation. \ourmethod{} has lower accuracy point
estimates in \textbf{10} conditions, ties in \textbf{3}, and improves in
\textbf{2}. Its macro accuracy difference relative to Initial is
\textbf{$-0.47$ points} (95\% cluster-bootstrap CI:
$[-2.24,1.31]$). The interval does not establish systematic harm, but
neither does this stress test support reliable transfer to non-additive
distortions. We therefore treat robustness to these transformations as
a boundary of the present training setup rather than extending the
additive-noise claim to all acoustic perturbations.

\begin{table*}[t]
    \centering
    \small
    \setlength{\tabcolsep}{5pt}
    \renewcommand{\arraystretch}{1.10}
    \caption{Non-additive stress test on 300 clean source tasks per condition.
    Values are accuracy percentages or \ourmethod{}-minus-Initial percentage
    points. Positive drop reduction favors \ourmethod{}.}
    \label{tab:app_nonadditive}
    \resizebox{0.94\textwidth}{!}{%
    \begin{tabular}{lrr|rrr|rr}
        \toprule
        & \multicolumn{2}{c|}{Mean accuracy} &
        \multicolumn{3}{c|}{$\Delta$ by severity} &
        \multicolumn{2}{c}{Mean robustness change} \\
        Distortion & Initial & Echo & Mild & Medium & Severe & $\Delta$Acc & Drop reduction \\
        \midrule
        Clean reference & 76.33 & 76.00 & -- & -- & -- & $-$0.33 & -- \\
        \midrule
        Bandwidth & 75.44 & 74.44 & $-$0.67 & $-$0.67 & $-$1.67 & $-$1.00 & $-$0.67 \\
        Channel filter & 75.78 & 75.33 & $-$0.33 & $-$0.67 & $-$0.33 & $-$0.44 & $-$0.11 \\
        Clipping & 76.44 & 76.33 & $-$0.33 & $-$0.67 & +0.67 & $-$0.11 & +0.22 \\
        Codec & 73.33 & 72.89 & $-$1.67 & 0.00 & +0.33 & $-$0.44 & $-$0.11 \\
        Reverberation & 75.78 & 75.44 & 0.00 & 0.00 & $-$1.00 & $-$0.33 & 0.00 \\
        \midrule
        \textbf{Macro over 15 conditions} & \textbf{75.36} & \textbf{74.89} &
        \multicolumn{3}{c|}{2 improved / 3 tied / 10 lower} &
        \textbf{$-$0.47} & \textbf{$-$0.13} \\
        \bottomrule
    \end{tabular}%
    }
    \vspace{0.2em}

    {\footnotesize Macro $\Delta$Acc 95\% cluster-bootstrap CI:
    $[-2.24,1.31]$.}
\end{table*}

\section{Additional Results for RQ3}
\label{app:rq3}

Clean--noisy agreement alone cannot establish reliance on acoustic
evidence: a model may answer consistently from the question or answer
choices. We therefore hold the prompt, choices, and decoding procedure
fixed while changing only the audio. The main intervention study uses
1,200 Qwen2.5-Omni examples at $-10\,\mathrm{dB}$ and evaluates
ground-truth task accuracy. Random supplies an unrelated waveform,
Shuffled permutes the audio--question pairing, and Silence removes the
waveform. These conditions assess whether the \emph{matched} acoustic
input contributes to a correct prediction, rather than treating
agreement itself as a grounding measure.

\subsection{Paired Audio-Control Transitions}
\label{app:rq3_transitions}

The main paper reports absolute accuracy under the four audio
conditions. Table~\ref{tab:app_audio_transitions} additionally tracks
paired changes in correctness. For \ourmethod{}, matched audio corrects
\textbf{138} cases that are wrong under Silence and breaks \textbf{90},
a net difference of \textbf{48} cases ($p=0.0018$). Relative to Random
audio, matched audio wins \textbf{129} cases and loses \textbf{52}
($p=9.75\times10^{-9}$). Initial also benefits from matched audio,
so these tests demonstrate that \ourmethod{}'s predictions use relevant
acoustic evidence; they do not establish that increased audio
dependence, by itself, explains the full post-training gain. The model-specific gap is computed on each checkpoint's own subset of
questions answered incorrectly under Silence. The shared-failed gap restricts
both checkpoints to the same 468 questions for which their recorded Silence
predictions are incorrect. Because these subsets are selected using observed
Silence outcomes, both estimates are descriptive conditional quantities rather
than unbiased population effects. 

\begin{table*}[t]
    \centering
    \scriptsize
    \setlength{\tabcolsep}{5pt}
    \renewcommand{\arraystretch}{1.12}
    \caption{Paired audio-control transitions on the 1,200-example diagnostic
    set at $-10\,\mathrm{dB}$. Fixed/Broken compares Matched with Silence;
    Matched/Random win compares correctness under the two audio assignments.
    Gaps are percentage points with paired 95\% bootstrap intervals.}
    \label{tab:app_audio_transitions}
    \resizebox{\textwidth}{!}{%
    \begin{tabular}{lrrr|rrr|ll}
        \toprule
        & \multicolumn{3}{c|}{Silence control} &
        \multicolumn{3}{c|}{Random-audio control} &
        \multicolumn{2}{c}{Matched--Random gap} \\
        \cmidrule(lr){2-4}\cmidrule(lr){5-7}\cmidrule(lr){8-9}
        Model & Fixed/Broken & Net & $p$ & Matched/Random win & Net & $p$ &
        Model-specific & Shared failed ($N=468$) \\
        \midrule
        Initial & 157/129 & +28 & 0.110 & 123/59 & +64 &
        $2.40{\times}10^{-6}$ & 10.00 [6.48, 13.33] & 10.68 [7.05, 14.32] \\
        \rowcolor{highlightblue}
        \textbf{\ourmethod{}} & 138/90 & +48 & 0.0018 & 129/52 & +77 &
        $9.75{\times}10^{-9}$ & 11.27 [7.79, 14.96] & 12.39 [8.76, 16.03] \\
        \midrule
        \ourmethod{} $-$ Initial & $-19/-39$ & +20 & -- & +6/$-7$ & +13 & -- & +1.27 & +1.71 \\
        \bottomrule
    \end{tabular}%
    }
\end{table*}

\begin{table*}[t]
    \centering
    \scriptsize
    \setlength{\tabcolsep}{4.2pt}
    \renewcommand{\arraystretch}{1.05}

    \caption{ECAC results on 470 preregistered eligible pairs.
    The upper table reports accuracy and changes from Initial.
    The lower table reports paired contrasts. Critical drop reduction is
    the comparator's clean-to-Critical drop minus \ourmethod{}'s; EPI is
    the Critical gain minus the Irrelevant gain.}
    \label{tab:app_ecac}

    \textbf{Panel A: Accuracy under different acoustic conditions}
    \vspace{0.3em}

    \begin{tabular*}{\textwidth}{
        @{\extracolsep{\fill}}
        lrrrrr
        @{}
    }
        \toprule
        & \multicolumn{5}{c}{Accuracy (change vs.\ Initial)} \\
        \cmidrule(lr){2-6}
        Method
        & Clean
        & Critical
        & Irrelevant
        & Global
        & Mean \\
        \midrule

        Initial
        & 72.13
        & 68.30
        & 68.72
        & 47.87
        & 64.26 \\

        Noisy-SFT
        & 68.09 {\scriptsize($-$4.04)}
        & 64.89 {\scriptsize($-$3.40)}
        & 64.89 {\scriptsize($-$3.83)}
        & 44.89 {\scriptsize($-$2.98)}
        & 60.69 {\scriptsize($-$3.56)} \\

        NoiseAug-SFT
        & 67.87 {\scriptsize($-$4.26)}
        & 64.47 {\scriptsize($-$3.83)}
        & 65.53 {\scriptsize($-$3.19)}
        & 44.68 {\scriptsize($-$3.19)}
        & 60.64 {\scriptsize($-$3.62)} \\

        Standard GRPO+KL
        & 68.94 {\scriptsize($-$3.19)}
        & 63.83 {\scriptsize($-$4.47)}
        & 65.74 {\scriptsize($-$2.98)}
        & 47.23 {\scriptsize($-$0.64)}
        & 61.44 {\scriptsize($-$2.82)} \\

        \ourmethod{}-StdRatio
        & 69.79 {\scriptsize($-$2.34)}
        & 65.32 {\scriptsize($-$2.98)}
        & 66.38 {\scriptsize($-$2.34)}
        & 47.02 {\scriptsize($-$0.85)}
        & 62.13 {\scriptsize($-$2.13)} \\

        \ourmethod{}-Ungated
        & 69.79 {\scriptsize($-$2.34)}
        & 65.96 {\scriptsize($-$2.34)}
        & 65.74 {\scriptsize($-$2.98)}
        & 47.02 {\scriptsize($-$0.85)}
        & 62.13 {\scriptsize($-$2.13)} \\

        \rowcolor{highlightblue}
        \textbf{\ourmethod{}}
        & \textbf{70.21} {\scriptsize($-$1.91)}
        & \textbf{68.09} {\scriptsize($-$0.21)}
        & \textbf{66.81} {\scriptsize($-$1.91)}
        & \textbf{49.36} {\scriptsize(+1.49)}
        & \textbf{63.62} {\scriptsize($-$0.64)} \\

        \bottomrule
    \end{tabular*}

    \vspace{1.2em}

    \textbf{Panel B: Paired statistical contrasts}
    \vspace{0.3em}

    \setlength{\tabcolsep}{3.7pt}
    \begin{tabular*}{\textwidth}{
        @{\extracolsep{\fill}}
        lrrrrr
        @{}
    }
        \toprule
        Contrast
        & Critical gain [95\% CI]
        & Global gain
        & Drop reduction [95\% CI]
        & EPI [95\% CI]
         \\
        \midrule

        \ourmethod{} $-$ Initial
        & $-$0.21 [$-$2.55, 2.13]
        & +1.49
        & +1.70 [$-$0.85, 4.47]
        & +1.70 [$-$0.64, 4.04]
        \\

        \rowcolor{highlightblue}
        \ourmethod{} $-$ Standard GRPO+KL
        & \textbf{+4.26 [1.91, 6.60]}
        & \textbf{+2.13}
        & \textbf{+2.98 [0.21, 5.74]}
        & \textbf{+3.19 [0.85, 5.53]}
         \\

        \bottomrule
    \end{tabular*}
\end{table*}

\subsection{Evidence-Critical Acoustic Corruption}
\label{app:rq3_ecac}

The evidence-critical acoustic corruption (ECAC) test probes a more
localized intervention. It compares matched-energy corruption of a
model-localized critical window, an irrelevant window, or the full
waveform. Table~\ref{tab:app_ecac} reports results on 470
preregistered eligible pairs. In Panel A, \ourmethod{} attains
\textbf{68.09\%} accuracy under critical-window corruption, compared
with \textbf{63.83\%} for Standard GRPO+KL. Panel B quantifies this
as a \textbf{+4.26-point} paired difference (95\% CI:
$[1.91,6.60]$). \ourmethod{} also reduces the clean-to-Critical
accuracy drop by \textbf{2.98 points} (95\% CI: $[0.21,5.74]$)
relative to that optimizer and improves the evidence preservation
index by \textbf{3.19 points} (95\% CI: $[0.85,5.53]$).

The comparison with Initial is more limited: \ourmethod{}'s Critical
accuracy is \textbf{$-0.21$ points} different (95\% CI:
$[-2.55,2.13]$), and its Clean accuracy in this subset is also below
Initial's. Thus, Table~\ref{tab:app_ecac} supports improved resistance
to evidence-critical corruption \emph{relative to the controlled
optimizer}; it does not show uniformly better acoustic grounding than
the initial checkpoint.

For a comparator $b$, we define
\begin{align}
\Delta_{\mathrm{crit}}(b)
&=
\mathrm{Acc}_{\mathrm{crit}}(\ourmethod{})
-\mathrm{Acc}_{\mathrm{crit}}(b), \\
\Delta_{\mathrm{irr}}(b)
&=
\mathrm{Acc}_{\mathrm{irr}}(\ourmethod{})
-\mathrm{Acc}_{\mathrm{irr}}(b), \\
\mathrm{EPI}(b)
&=
\Delta_{\mathrm{crit}}(b)-\Delta_{\mathrm{irr}}(b), \\
\mathrm{DropRed.}(b)
&=
\left[
\mathrm{Acc}_{\mathrm{clean}}(b)
-\mathrm{Acc}_{\mathrm{crit}}(b)
\right]
-
\left[
\mathrm{Acc}_{\mathrm{clean}}(\ourmethod{})
-\mathrm{Acc}_{\mathrm{crit}}(\ourmethod{})
\right].
\end{align}
Positive EPI and drop reduction indicate better preservation of
evidence-critical information by \ourmethod{} relative to comparator $b$.

\begin{table*}[t]
    \centering
    \small
    \setlength{\tabcolsep}{4.5pt}
    \renewcommand{\arraystretch}{1.06}
    \caption{Teacher-reliability audit on 3,596 paired examples at
    $-10\,\mathrm{dB}$. Correct/Wrong denotes teacher correctness, high/low
    denotes the audit's confidence partition, WT imitation is agreement with
    an incorrect teacher, and WT recovery is accuracy when that teacher is
    wrong. All values are percentages.}
    \label{tab:app_teacher_reliability}
    \resizebox{\textwidth}{!}{%
    \begin{tabular}{lrrrrrrr}
        \toprule
        Method & Overall & Correct-high & Correct-low & Wrong-high & Wrong-low & WT imitation & WT recovery \\
        \midrule
        Initial & 59.40 & 79.50 & 67.17 & 42.09 & 39.16 & 33.87 & 39.80 \\
        Noisy-SFT & 60.37 & 79.50 & 65.81 & 39.24 & 44.62 & 31.08 & 43.44 \\
        NoiseAug-SFT & 60.07 & 79.50 & 65.50 & 38.61 & 44.18 & 31.15 & 42.95 \\
        Standard GRPO+KL & 61.54 & 81.80 & 67.65 & 41.46 & 43.82 & 32.68 & \textbf{43.30} \\
        \ourmethod{}-StdRatio & 61.04 & 82.46 & 69.01 & 40.82 & 40.32 & 35.06 & 40.43 \\
        \ourmethod{}-Ungated & 61.10 & 82.57 & 68.93 & 40.51 & 40.59 & 34.85 & 40.57 \\
        \rowcolor{highlightblue}
        \textbf{\ourmethod{}} & \textbf{62.99} & \textbf{84.76} & \textbf{71.01} & \textbf{43.04} & 41.85 & 34.15 & 42.11 \\
        \bottomrule
    \end{tabular}%
    }
\end{table*}

\begin{table}[t]
    \centering
    \small
    \setlength{\tabcolsep}{5pt}
    \renewcommand{\arraystretch}{1.08}
    \caption{Task-positive gate audit. All values are percentages.
Verified-wrong is a conservative lower bound because unresolved free-form
responses are not labeled incorrect.}
    \label{tab:app_gate_audit}
    \resizebox{0.88\linewidth}{!}{%
    \begin{tabular}{lrrr}
        \toprule
        Method & Positive & Verified-wrong/positive & Wrong-teacher blocked \\
        \midrule
        Standard GRPO+KL & 52.84 & 1.62 & 98.01 \\
        \ourmethod{}-StdRatio & 54.16 & 2.50 & 97.28 \\
        \rowcolor{highlightblue}
        \textbf{\ourmethod{}} & \textbf{56.33} & 2.09 & 97.79 \\
        \bottomrule
    \end{tabular}%
    }
\end{table}

\begin{table*}[t]
    \centering
    \scriptsize
    \setlength{\tabcolsep}{4.0pt}
    \renewcommand{\arraystretch}{1.06}
    \caption{Auxiliary open-ended evaluation on 300 AIR-Bench Chat examples at
    $-10\,\mathrm{dB}$. Rates come from an audio-conditioned automatic judge.
    The last two columns are \ourmethod{}-minus-comparator paired differences
    with 95\% bootstrap confidence intervals.}
    \label{tab:app_open_ended}
    \resizebox{\textwidth}{!}{%
    \begin{tabular}{lrrrrll}
        \toprule
        Model & Valid & Noisy correct & Noisy relevant & Noisy UAC $\downarrow$ & $\Delta$ correct & $\Delta$ relevant \\
        \midrule
        Initial & 100.00 & 36.00 & 48.00 & 17.00 & +2.00 [$-$4.33, 8.33] & +7.67 [1.33, 14.33] \\
        Noisy-SFT & 100.00 & 35.00 & 48.33 & 16.33 & +3.00 [$-$3.00, 9.00] & +7.33 [1.00, 13.67] \\
        NoiseAug-SFT & 100.00 & 35.00 & 48.33 & 17.67 & +3.00 [$-$3.33, 9.00] & +7.33 [0.67, 14.00] \\
        Standard GRPO+KL & 100.00 & 34.33 & 48.00 & \textbf{11.33} & +3.67 [$-$2.33, 10.00] & +7.67 [0.67, 14.67] \\
        \ourmethod{}-StdRatio & 100.00 & 34.33 & 47.33 & 12.67 & +3.67 [$-$2.33, 10.00] & +8.33 [1.33, 15.33] \\
        \ourmethod{}-Ungated & 100.00 & 36.67 & 49.00 & 16.00 & +1.33 [$-$4.67, 7.33] & +6.67 [0.00, 13.33] \\
        \rowcolor{highlightblue}
        \textbf{\ourmethod{}} & 100.00 & \textbf{38.00} & \textbf{55.67} & \textbf{11.33} & -- & -- \\
        \bottomrule
    \end{tabular}%
    }
    \vspace{0.2em}

    {\footnotesize \ourmethod{} Matched$-$Random correctness:
    $+12.00$ [5.67, 18.67]; Matched$-$Silence: $+13.67$ [7.00, 20.67].}
\end{table*}

\begin{table*}[t]
    \centering
    \scriptsize
    \setlength{\tabcolsep}{3.5pt}
    \renewcommand{\arraystretch}{1.06}
    \caption{Answer-option position audit at $-10\,\mathrm{dB}$ on 120 source
    questions under the original order and three cyclic rotations. Mean
    accuracy averages all orders; All-4 requires correctness in every order;
    Consistency requires the same answer content in every order.}
    \label{tab:app_option_position}
    \resizebox{\textwidth}{!}{%
    \begin{tabular}{lrrrrrrrr}
        \toprule
        Model & Original & Rot. 1 & Rot. 2 & Rot. 3 & Mean Acc. & All-4 & Consistency & Valid \\
        \midrule
        Initial & 69.17 & 65.00 & 70.83 & 65.83 & 67.71 & 53.33 & 62.50 & 100.00 \\
        Noisy-SFT & 69.17 & 61.67 & 61.67 & 64.17 & 64.17 & 49.17 & 54.17 & 100.00 \\
        NoiseAug-SFT & 70.00 & 60.83 & 62.50 & 64.17 & 64.38 & 49.17 & 53.33 & 100.00 \\
        Standard GRPO+KL & 71.67 & 63.33 & 65.83 & 64.17 & 66.25 & 50.83 & 55.83 & 100.00 \\
        \ourmethod{}-StdRatio & 69.17 & 64.17 & 70.00 & 66.67 & 67.50 & 52.50 & 60.83 & 100.00 \\
        \ourmethod{}-Ungated & 69.17 & 63.33 & 70.00 & 66.67 & 67.29 & 52.50 & 60.00 & 100.00 \\
        \rowcolor{highlightblue}
        \textbf{\ourmethod{}} & \textbf{72.50} & \textbf{66.67} & \textbf{70.83} & \textbf{68.33} & \textbf{69.58} & \textbf{54.17} & 61.67 & 100.00 \\
        \midrule
        \multicolumn{3}{l}{\ourmethod{} $-$ Initial} &
        \multicolumn{2}{c}{Mean: +1.88 [$-$1.67, 5.42]} &
        \multicolumn{2}{c}{All-4: +0.83 [$-$5.83, 7.50]} &
        \multicolumn{2}{c}{Consistency: $-$0.83 [$-$8.33, 6.67]} \\
        \multicolumn{3}{l}{\ourmethod{} $-$ Standard GRPO+KL} &
        \multicolumn{2}{c}{Mean: +3.33 [0.00, 6.67]} &
        \multicolumn{2}{c}{All-4: +3.33 [$-$3.33, 10.00]} &
        \multicolumn{2}{c}{Consistency: +5.83 [$-$2.50, 14.17]} \\
        \bottomrule
    \end{tabular}%
    }
\end{table*}

\subsection{Teacher Reliability and Reward Gating}
\label{app:rq3_teacher}

The clean-conditioned teacher provides useful guidance but can also
make incorrect predictions. Table~\ref{tab:app_teacher_reliability}
stratifies 3,596 paired examples by teacher correctness and the audit's
confidence partition. \ourmethod{} reaches \textbf{62.99\%} overall
accuracy, compared with \textbf{61.54\%} for Standard GRPO+KL. Its
accuracy is higher in both correct-teacher partitions
(\textbf{84.76\%} versus \textbf{81.80\%} for Correct-high, and
\textbf{71.01\%} versus \textbf{67.65\%} for Correct-low).
The pattern is less uniform when the teacher is wrong:
\ourmethod{}'s wrong-teacher recovery is \textbf{42.11\%}, below
Standard GRPO+KL's \textbf{43.30\%}. These descriptive strata
therefore do not justify a claim that \ourmethod{} reliably corrects
teacher errors.

Importantly, \ourmethod{}'s wrong-teacher imitation rate is
\textbf{34.15\%}, while its task accuracy on wrong-teacher examples is
\textbf{42.11\%}; it does not deterministically reproduce the
teacher's answer. Table~\ref{tab:app_gate_audit} further shows that
the task-positive gate excludes \textbf{97.79\%} of
\emph{verified-wrong} teacher signals from the reward. Because
unresolved free-form answers are not certified as wrong, this is an
audit of identifiable errors, not a guarantee that all erroneous
teacher guidance is removed. The confidence partitions are likewise
descriptive and should not be interpreted as calibrated selective-risk
estimates.

\begin{table*}[t]
    \centering
    \small
    \setlength{\tabcolsep}{6pt}
    \renewcommand{\arraystretch}{1.06}
    \caption{Independent noise-resampling retest at $-10\,\mathrm{dB}$. Both
    methods receive the same 1,200 waveforms. Gain is \ourmethod{} minus
    Standard GRPO+KL; intervals use 10,000 paired bootstrap resamples
    clustered by clean source. C/B denotes corrected/broken cases.}
    \label{tab:app_noise_resampling}
    \begin{tabular*}{\textwidth}{@{\extracolsep{\fill}}lrrrrrcr}
        \toprule
        Setting & $N$ & Sources & GRPO+KL & Echo & Gain & 95\% CI & C/B \\
        \midrule
        \multicolumn{8}{l}{\textit{Independent realizations}} \\
        Realization 1 & 600 & 600 & 60.67 & 61.67 & +1.00 & [$-$2.00, 4.00] & 45/39 \\
        Realization 2 & 600 & 600 & 60.17 & 61.50 & +1.33 & [$-$1.67, 4.17] & 44/36 \\
        \midrule
        \multicolumn{8}{l}{\textit{Prespecified pooled analysis}} \\
        \textbf{Overall} & 1,200 & 600 & 60.42 & \textbf{61.58} & +1.17 & [$-$1.33, 3.67] & 89/75 \\
        Music & 400 & 200 & 60.25 & 63.50 & +3.25 & [$-$1.25, 8.00] & 34/21 \\
        Sound & 400 & 200 & 63.75 & 63.50 & $-$0.25 & [$-$4.50, 4.00] & 24/25 \\
        Speech & 400 & 200 & 57.25 & 57.75 & +0.50 & [$-$3.75, 4.75] & 31/29 \\
        \bottomrule
    \end{tabular*}
    
\end{table*}

\subsection{Open-Ended and Option-Position Controls}
\label{app:rq3_shortcuts}

Table~\ref{tab:app_open_ended} extends the audio intervention to
300 AIR-Bench Chat examples evaluated by an audio-conditioned
automatic judge. For \ourmethod{}, matched audio increases judged
correctness over Random by \textbf{12.00 points} (95\% CI:
$[5.67,18.67]$) and over Silence by \textbf{13.67 points}
(95\% CI: $[7.00,20.67]$). However, its matched-audio correctness
exceeds Initial by only \textbf{2.00 points}, with a confidence
interval crossing zero ($[-4.33,8.33]$). This auxiliary, automatically
judged analysis supports sensitivity to matched audio; it is not an
official AIR-Bench score or evidence of a reliable cross-model
accuracy gain. \ourmethod{} also reaches \textbf{55.67\%} judged relevance, a
\textbf{+7.67}-point difference from Initial
(95\% CI \textbf{[1.33, 14.33]}), while its UAC rate of
\textbf{11.33\%} ties Standard GRPO+KL. Thus, the auxiliary advantage is
clearest for judged relevance rather than uniform across all judge
dimensions.

Table~\ref{tab:app_option_position} examines a different shortcut:
dependence on answer-option position. Across the original order and
three cyclic rotations of 120 source questions, \ourmethod{}'s mean
accuracy exceeds Initial's by \textbf{1.88 points} (95\% CI:
$[-1.67,5.42]$), while answer-content consistency differs by
\textbf{$-0.83$ points} (95\% CI: $[-8.33,6.67]$). Neither interval
establishes a reliable change. The option-order test therefore does
not show that this shortcut explains the matched-audio effect, but
it also does not establish improved option-order robustness. Relative to Standard GRPO+KL, the mean-accuracy difference is
\textbf{+3.33} points, but its interval touches zero
(\textbf{[0.00, 6.67]}), while the All-4 and consistency intervals both cross
zero. We therefore treat all option-position comparisons as inconclusive.

\subsection{Independent Noise Resampling}
\label{app:rq3_resampling}

Table~\ref{tab:app_noise_resampling} tests whether the comparison with
Standard GRPO+KL depends on a particular noise realization. We hold
600 clean source tasks fixed and generate two new noise waveforms per
source; both methods receive the same waveform in each comparison.
\ourmethod{}'s accuracy differences are \textbf{+1.00 point}
(95\% CI: $[-2.00,4.00]$) and \textbf{+1.33 points}
(95\% CI: $[-1.67,4.17]$) on the two realizations. The prespecified
pooled analysis covers 1,200 waveforms clustered by the 600 clean
sources and yields \textbf{+1.17 points} (95\% CI:
$[-1.33,3.67]$), corresponding to \textbf{89} corrected and
\textbf{75} broken predictions relative to Standard GRPO+KL.
The point-estimate direction repeats across these two draws, but all
reported intervals include zero; Table~\ref{tab:app_noise_resampling}
does not establish a statistically reliable resampled advantage.

\paragraph{Summary of the RQ3 evidence.}
The matched-audio controls, including the auxiliary open-ended test,
show that question-relevant audio contributes to \ourmethod{}'s
predictions. ECAC indicates an advantage over the controlled optimizer
under localized evidence corruption, while the teacher audit shows
both the value and the limits of clean-conditioned guidance. At the
same time, performance under mismatched audio remains nontrivial,
ECAC does not establish an advantage over Initial, and the independent
noise-resampling interval includes zero. These analyses support a
bounded claim of acoustic-evidence dependence under the tested
interventions, not a claim that every gain is caused by stronger
grounding.

\begin{wrapfigure}{r}{0.50\textwidth}
    \centering
     \vspace{-5.0em}
    \includegraphics[width=\linewidth]{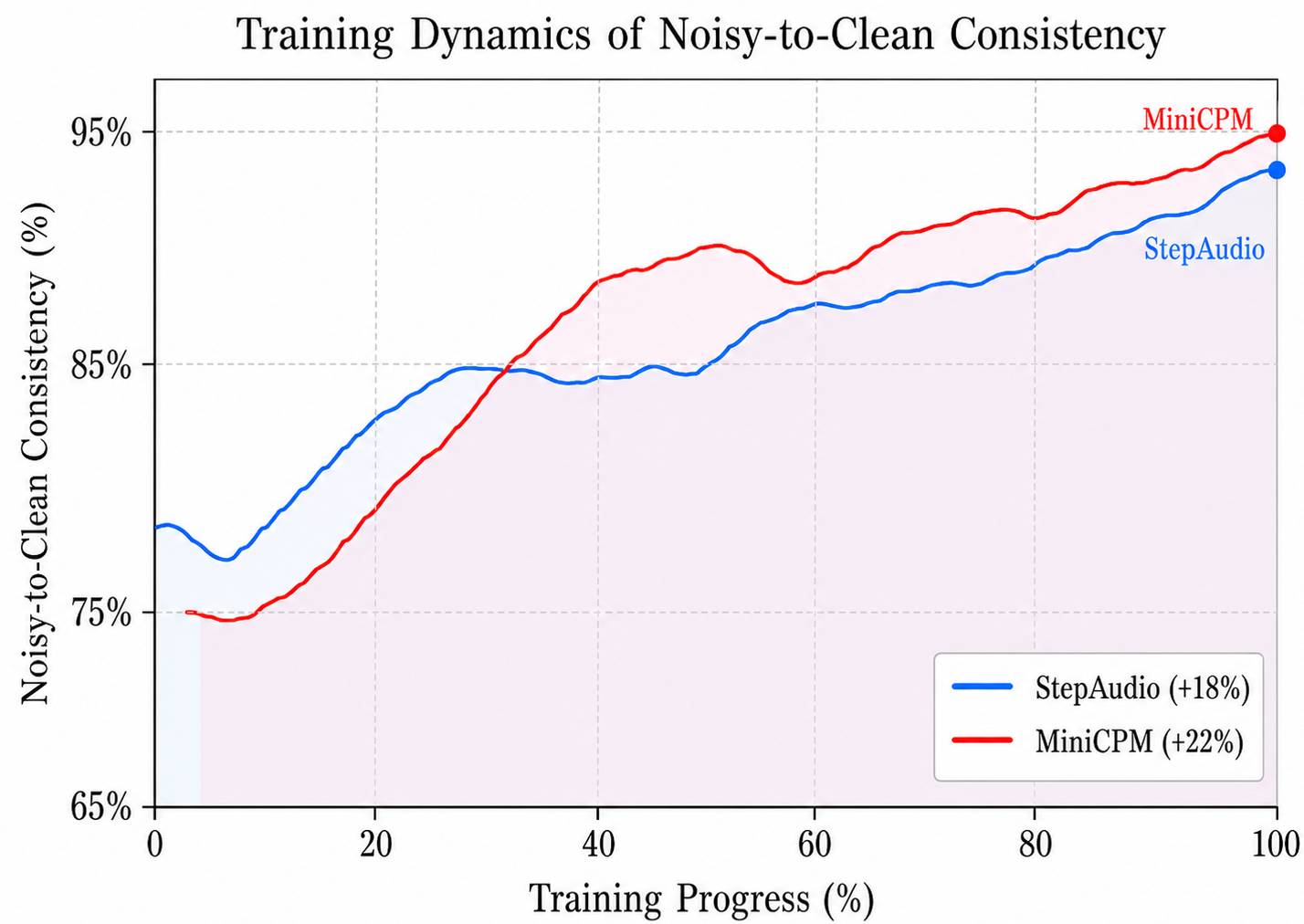}
    \vspace{-0.8em}
    \caption{Clean--noisy consistency over post-training progress.
    MiniCPM-o improves rapidly during the early stage, whereas Step-Audio2
    exhibits a steadier increase.}
    \label{fig:app_training_dynamics}
      \vspace{-0.5\baselineskip}
\end{wrapfigure}
\subsection{Optimization Dynamics}
\label{app:optimization_dynamics}

Figure~\ref{fig:app_training_dynamics} reports clean--noisy prediction
consistency during post-training. The curves show an overall upward trend in clean--noisy consistency during
post-training, although the trajectories are not monotonic. Because
consistency does not imply correctness, this figure is used only as an
optimization diagnostic; task accuracy is evaluated separately.

\end{document}